\documentclass[10pt,twocolumn,letterpaper]{article}

\usepackage[pagenumbers]{cvpr} 

\usepackage{graphicx}
\usepackage{amsmath}
\usepackage{amssymb}
\usepackage{booktabs}

\usepackage{algorithm}
\usepackage{tabularx}
\usepackage{threeparttable}
\usepackage{booktabs}
\usepackage{multirow}
\usepackage[table]{xcolor}
\usepackage{algpseudocode}
\usepackage{lineno}
\usepackage{makecell}
\newcommand{\down}[1]{\tiny{#1}}

\usepackage[pagebackref,breaklinks,colorlinks]{hyperref}

\usepackage[capitalize]{cleveref}
\crefname{section}{Sec.}{Secs.}
\Crefname{section}{Section}{Sections}
\Crefname{table}{Table}{Tables}
\crefname{table}{Tab.}{Tabs.}

\usepackage[accsupp]{axessibility}  

\def\cvprPaperID{7826} 
\def\confName{CVPR}
\def\confYear{2023}

\begin{document}

\title{Master: Meta Style Transformer for Controllable Zero-Shot and Few-Shot Artistic Style Transfer}

\author{
Hao Tang$^{1}$\footnotemark[1] ~~~~Songhua Liu$^{2}$\footnotemark[1]~ ~~~~Tianwei Lin$^{3}$  ~~~~Shaoli Huang$^{4}$  \\
~~~~Fu Li$^{3}$  ~~~~Dongliang He$^{3}$  ~~~~Xinchao Wang$^{2}\footnotemark[2]  $    \\
{\normalsize
{$^1$}Center for Data Science, Peking University ~~{$^2$}National University of Singapore}\\
{\normalsize{\hspace*{-16pt}}
~~{$^3$}VIS, Baidu Inc.
~~{$^4$}Tencent AI Lab
}
\\
{\tt\small tanghao@stu.pku.edu.cn
~~songhua.liu@u.nus.edu}\\
{\tt\small\{lintianwei01,lifu,hedongliang01\}@baidu.com
~~shaolihuang@tencent.com
~~xinchao@nus.edu.sg
}
}

\twocolumn[{%

\maketitle

\begin{figure}[H]
\vspace{-1.2cm}
\hsize=\textwidth 
\centering
\includegraphics[width=\textwidth]{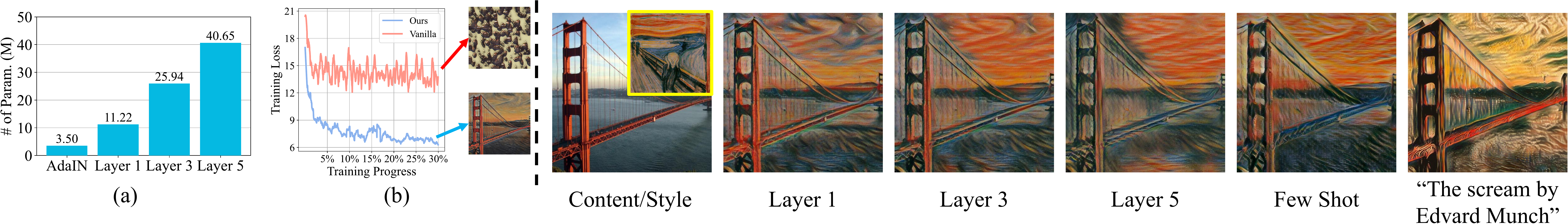}
\vspace{-0.7cm}
\caption{Left: With more Transformer layers, the number of parameters and training difficulty increase significantly. Right: Our method supports test-time controlling of stylization degree via tuning the number of adopted Transformer layers. It is also readily adaptable to few-shot style transfer, where stylization with only 1 layer can be further improved. Text-guided few-shot style transfer is achievable.}
  \label{fig:1}
\end{figure}
}]
\renewcommand{\thefootnote}{\fnsymbol{footnote}}
\footnotetext[1]{Equal contribution.}
\footnotetext[2]{Corresponding author.}

\begin{abstract}
\vspace{-0.3cm}
Transformer-based models achieve favorable performance in artistic style transfer recently thanks to its global receptive field and powerful multi-head/layer attention operations. 
Nevertheless, the over-paramerized multi-layer structure increases parameters significantly and thus presents a heavy burden for training. 
Moreover, for the task of style transfer, vanilla Transformer that fuses content and style features by residual connections is prone to content-wise distortion. 
In this paper, we devise a novel Transformer model termed as \emph{Master} specifically for style transfer. 
On the one hand, in the proposed model, different Transformer layers share a common group of parameters, which (1) reduces the total number of parameters, (2) leads to more robust training convergence, and (3) is readily to control the degree of stylization via tuning the number of stacked layers freely during inference. 
On the other hand, different from the vanilla version, we adopt a learnable scaling operation on content features before content-style feature interaction, which better preserves the original similarity between a pair of content features while ensuring the stylization quality. 
We also propose a novel meta learning scheme for the proposed model so that it can not only work in the typical setting of arbitrary style transfer, but also adaptable to the few-shot setting, by only fine-tuning the Transformer encoder layer in the few-shot stage for one specific style. 
Text-guided few-shot style transfer is firstly achieved with the proposed framework. 
Extensive experiments demonstrate the superiority of Master under both zero-shot and few-shot style transfer settings. 
\end{abstract}

\section{Introduction}
\label{sec:intro}

Artistic style transfer aims at applying style patterns like colors and textures of a reference image to a given content image while preserving the semantic structure of the content. 
In contrast to the pioneering optimization method~\cite{Gatys_2016_CVPR} and early per-style-per-model methods like \cite{johnson2016perceptual,liu2020stable}, arbitrary style transfer methods~\cite{li2017universal,huang2017arbitrary,park2019arbitrary,li2018learning,deng2020arbitrary1,Liu_2021_ICCV,jing2022learning} enable real time style transfer for any style image in the test time in a zero-shot manner.
The flexibility has led to this \emph{arbitrary-style-per-model} fashion to dominate style transfer research.

Recently, to enhance the representation of global information in arbitrary style transfer, Transformer~\cite{vaswani2017attention} is introduced to this area~\cite{deng2021stytr2}, leveraging the global receptive field and powerful multi-head/layer structure, and achieves superior performance. 
Nevertheless, the over-parameterized multi-layer structure increases model parameters significantly. 
As shown in Fig. \ref{fig:1}(a), there are 25.94M learnable parameters for a 3-layer Transformer structure in StyTr2~\cite{deng2021stytr2}, v.s. 3.50M in AdaIN~\cite{huang2017arbitrary}, a simple but effective baseline in arbitrary style transfer. 
Such a large model for standard Transformer inevitably presents a heavy burden for training. 
As shown in Fig. \ref{fig:1}(b), when there are more than 4 layers, vanilla Transformer even fails to get convergent in training, which limits the scalability of the Transformer model in style transfer. 

\begin{figure}
\centering
  \includegraphics[width=\linewidth]{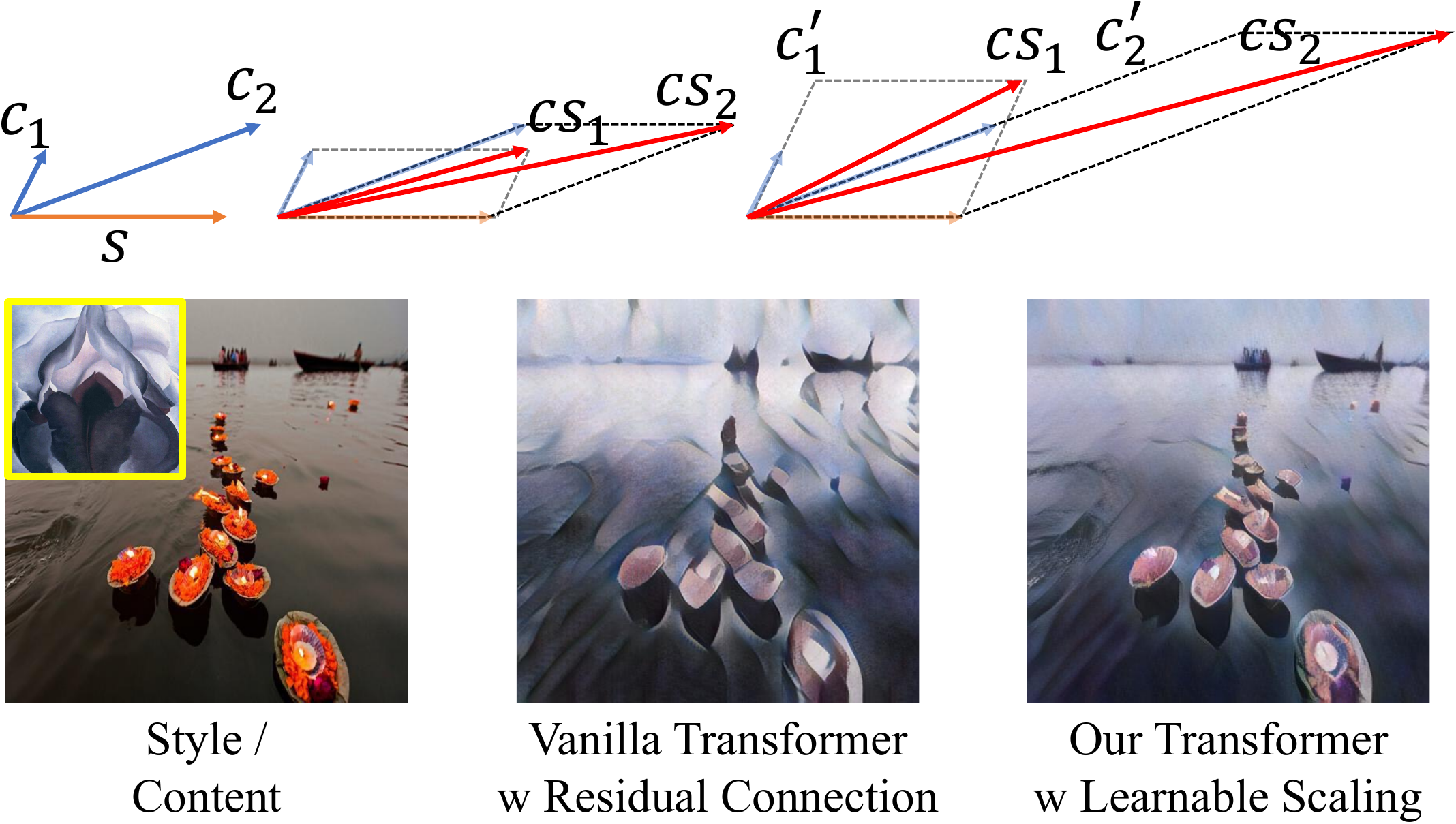}
  \caption{Residual connection in the vanilla Transformer tends to destroy the original similarity relationship on content structure in style transfer task. Our model is designed to address this problem with learnable scaling parameters. The top row shows a simple 2D visualization and the bottom one provides a qualitative example.
  }
  \vspace{-0.5cm}
  \label{fig:motivation}
\end{figure}

Moreover, vanilla Transformer relies on residual connections~\cite{he2016deep} to stylize content features, which suffers from the content-distortion problem. 
We illustrate this effect with a 2D visualization in Fig. \ref{fig:motivation}(top), where residual connections lead the transformation results of two content feature vectors to move towards the dominated style features and thereby tend to eliminate their original distinction. 
The visual effect is that the stylized images would be dominated by some strong style features, such as salient edges, with the original self-(dis)similarity of content structures destroyed, as the example shown in Fig.~\ref{fig:motivation}(bottom). 

Focusing on these drawbacks, in this paper, we are dedicated to devising a novel Transformer architecture specifically for artistic style transfer. 
On the one hand, in the proposed model, different Transformer layers share a common group of parameters and a random number of stacked layers are adopted for each training iteration. 
Compared with the original version, sharing parameters across different layers reduces the total number of parameters significantly and leads to more convenient training convergence. 
As a byproduct, it is also readily for our model to control the degree of stylization via tuning the number of stacked layers freely in the inference time, as shown in Fig. \ref{fig:1}(right). 
On the other hand, we equip Transformer with learnable scale parameters for content-style interactions instead of residual connections, which alleviates content distortion to a large degree and better preserves content structures while rendering vivid style patterns simultaneously, as shown in the 2D visualization and the qualitative example in Fig. \ref{fig:motivation}. 

Furthermore, beyond the typical zero-shot arbitrary style transfer, leveraging a meta learning scheme, our method is adaptable to the few-shot setting. 
By only fine-tuning the Transformer encoder layer in the few-shot stage, rapid adaptation to the model for a specific style within a limited number of updates is possible, where the stylization with only 1 layer can be further improved, as shown in Fig. \ref{fig:1}. 
Beyond that, we first achieve text-guided few-shot style transfer with this framework, which largely alleviates the training burden of previous per-text-per-model solution. 
In this sense, we term the overall pipeline \emph{\textbf{M}et\textbf{a} \textbf{S}tyle \textbf{T}ransform\textbf{er}} (\emph{Master}). 
Our contributions are summarized as follows:
\begin{itemize}
\item We propose a novel Transformer architecture specifically for artistic style transfer. It shares parameters between different layers, which not only helps training convergence, 
but also allows convenient control over the stylization effect.
\item We identity the content distortion problem of residual connections in Transformer and propose learnable scale parameters as an option to alleviate the problem. 
\item We introduce a meta learning framework for adapting original training setting of zero-shot style transfer to the few-shot scenario, with which our Master achieves very good trade-off between flexibility and quality. 
\item Experiments show that our model achieves 
results better than those of
arbitrary-style-per-model methods.
Furthermore, under the few-shot setting, either conditioned on image or text, Master can even yield performance on par with that of per-style-per-model methods with significantly less training cost.
\end{itemize}

\section{Related Works}

\begin{figure*}
\centering
  \includegraphics[width=\textwidth]{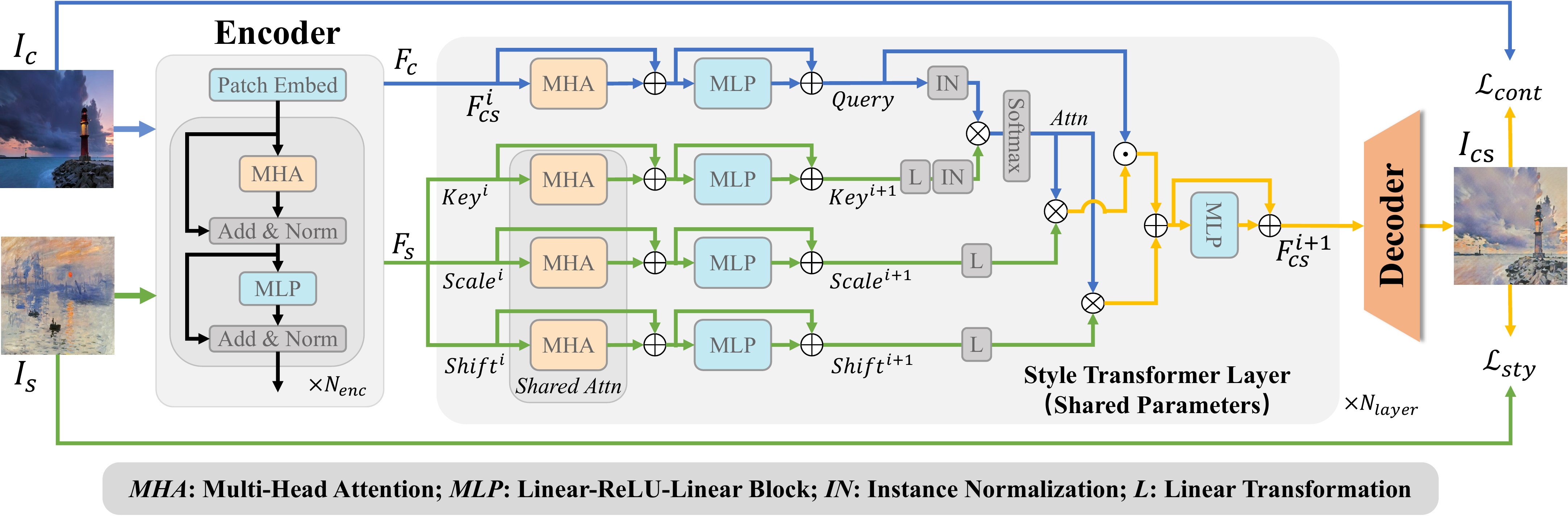}
  \caption{Overview of our model architecture.}
  \vspace{-0.5cm}
  \label{fig:framework}
\end{figure*}

\noindent
\textbf{Arbitrary-Style-Per-Model Methods.} 
A lot of works achieve arbitrary style transfer via global feature transformation, \textit{e.g.}, WCT~\cite{li2017universal}, AdaIN~\cite{huang2017arbitrary}, Linear style transfer~\cite{li2018learning}, DIN~\cite{jing2020dynamic}, MCCNet~\cite{deng2020arbitrary2}, and MAST~\cite{huo2021manifold}. 
In general, they can achieve the most attractable style transfer speed but dismiss stylized effects for local details a lot. 
In order to add more consideration of local details to global transformation based methods, there are patch-similarity based solution~\cite{chen2016fast,gu2018arbitrary,sheng2018avatar} and attention based method~\cite{park2019arbitrary,yao2019attention,deng2020arbitrary1,Liu_2021_ICCV}. 
While paying  attention to local details,
it is still hard to transfer complex style patterns and prone to unsatisfactory distortions due to the simple swap and fusion strategies. 

\noindent
\textbf{Transformer in Style Transfer.} 
Transformer proposed in \cite{vaswani2017attention} is widely used in the natural language processing and has become a powerful baseline.
Recently, Transformer receives better performance than CNN models in many vision tasks and are involved in various model zoos~\cite{khan2021transformers,han2022survey,yang2022deep,yang2022KF,FangPrunging2023CVPR,Ye2023CVPR}. 
Specifically, in the field of style transfer, \cite{wu2021styleformer} propose StyFormer, where the transformation is driven by the cross-attention module in Transformer. 
\cite{deng2021stytr2} develop a Transformer model based on ViT~\cite{dosovitskiy2020image}. 
However, the number of parameters and training difficulty would increase considerably with the increasing of Transformer layers. 
Also, following residual connections in original Transformer, it suffers from the distortion problem on content structures as shown in Fig.~\ref{fig:motivation}. 
Similar problems also exist in \cite{LiuStyle2022ECCV}.

\noindent
\textbf{Meta Learning in Style Transfer.} 
On this routine, \cite{shen2017meta} train a meta network to predict parameters of the generator model for one style reference image. 
\cite{zhang2019metastyle} employ MAML algorithm~\cite{finn2017modelagnostic} to find a style-free model for fast adaptation.
Our method is different from theirs in three aspects: (1) our framework works for both few-shot and arbitrary style transfer thanks to the cross-attention in Transformer; 
(2) our meta learning algorithm is based on Reptile~\cite{nichol2018firstorder}, a first-order meta learning algorithm without the necessity to operate gradients of higher levels, which is more efficient in training; 
and (3) only style encoder instead of the whole model needs to be updated during the few-shot learning stage, which makes it more convenient in practice. 

\noindent
\textbf{Text-Guided Synthesis.}  
The emergence of CLIP model~\cite{radford2021learning} bridges text and image domain, which supports a series of works on text-guided synthesis~\cite{patashnik2021styleclip,gal2022stylegan,rombach2021highresolution}. 
In the field of style transfer, CLIPstyler~\cite{kwon2022clipstyler} proposes a patch-wise CLIP loss and achieves text-guided style transfer. 
Nevertheless, as a per-text-per-model solution, it is inconvenient in practice to handle a large number of text inputs. Despite the recent dataset distillation scheme may alleviate this issue,
existing approaches, however, largely focus on images as input~\cite{yu2023dataset,liu2022dataset,LiuSonghua2023CVPR}.
In this paper, we first achieve text-guided few-shot style transfer with the proposed meta learning algorithm, which improves the flexibility of CLIPstyler significantly. 

\section{Methods}

In this section, we give details of the proposed 
Meta Style Transformer (Master) for zero-shot and few-shot style transfer.
We first introduce the network architecture of the proposed model,
then illustrate how we train our model in a meta-learning fashion,
and finally describe loss function. 

\subsection{Network Architecture}
The proposed model comprises an encoder, a feature modification module, and a decoder, as demonstrated in Fig. \ref{fig:framework}. 
We employ the first 2 stages of Swin Transformer~\cite{liu2021swin} as the encoder ${\rm Enc}$ to extract common image features for both content and style images. 
The decoder ${\rm Dec}$ follows the setting of \cite{huang2017arbitrary} with 3 upsampling convolutional blocks. 
For feature modification, we propose a \emph{Style Transformer} module for transferring complex style patterns, which will be introduced later.
Taking a style image $I_s$ and a content image $I_c$ as input, we first divide them into $4\times4$ patches and extract their corresponding feature maps $F_c$ and $F_s$ with the Swin encoder ${\rm Enc}$:
\begin{equation}
\begin{aligned}
    F_c={\rm Enc}(I_c),\\
    F_s={\rm Enc}(I_s), \label{eq_enc}
\end{aligned}
\end{equation}
where the spatial scales for $F_c$ and $F_s$ are 1/8 of those for $I_c$ and $I_s$. 
Then, the embedded feature $F_{cs}$ is derived by:
\begin{equation}
    F_{cs}={\rm StyleTrans}(F_s,F_c), \label{eq_trans}
\end{equation}
where ${\rm StyleTrans}$ denotes the Style Transformer module, and $F_{cs}$ has the same shape as $F_c$. 
Finally, we can synthesize the stylized image $I_{cs}$ with the decoder as:
\begin{equation}
    I_{cs}={\rm Dec}(F_{cs}). \label{eq_dec}
\end{equation}

\noindent
\textbf{Style Transformer.} 
Similar to previous Transformer-based style transfer models like StyTr2~\cite{deng2021stytr2}, the Transformer encoder is used to encode style information while the Transformer decoder takes charge of content-style interaction. 
In our model, there is only one copy of parameters shared by all Transformer encoder and decoder layers. 
Also, different from standard Transformer where decoder layers would follow all encoder layers, encoder and decoder layers would be executed in an alternate fashion. 
The next layer would take the output of the current layer as input. 

Our Transformer decoder layer is composed of self-attention, cross-attention, and non-linear blocks, for content encoding and content-style interaction. 
Specifically, content features are first processed with a self-attention step. 
Then, cross-attention is conducted by taking the content encoding as query and the style encoding as key and value, followed by a MLP for non-linear transformation. 

Notably, in vanilla Transformer, features before and after the cross-attention step are fused by a residual connection, which is harmful for content structures as analyzed in Fig. \ref{fig:motivation}. 
We thereby replace the residual connection with dynamic and learnable scaling and shifting steps, whose parameters are determined by the style encoder. 
In this way, the output of style encoder should consist of three parts: key for the following cross-attention $K_{s}$, scaling parameters $V_{\sigma}$, and shifting parameters $V_{\mu}$. 
The prediction of each part shares a same self-attention map to save memory but uses independent non-linear transformation. 
The process in the Transformer encoder layer can be formulated as:
\begin{equation}
\begin{aligned}
    {\rm MHA}(Q,K,V)&=[head_1,...,head_h]W^O,\\
    head_i&={\rm Att}(QW^{Q_i},KW^{K_i},VW^{V_i}),\\
    {\rm Att}(Q,K,V)&={\rm softmax}(\frac{QK^{\top}}{\sqrt{d_k}}V),\\
    K'_{s}&=K_{s}+{\rm MHA}(K_{s},K_{s},K_{s}),\\
    K_{s}&=K'_{s}+{\rm MLP}(K'_{s}),\\
    V'_{\sigma}&=V_{\sigma}+{\rm MHA}(K_{s},K_{s},V_{\sigma}),\\
    V_{\sigma}&=V'_{\sigma}+{\rm MLP}(V'_{\sigma}),\\
    V'_{\mu}&=V_{\mu}+{\rm MHA}(K_{s},K_{s},V_{\mu}),\\
    V_{\mu}&=V'_{\mu}+{\rm MLP}(V'_{\mu}),\label{eq:encoder}
\end{aligned}
\end{equation}
where $K_{s}$, $V_{\sigma}$, and $V_{\mu}$ are initialized as the style feature $F_s$ before the first Transformer encoder layer. 
We do not incorporate normalization to encode style features since second-order statistics can largely represent style information. 

Then, in the cross-attention of Transformer decoder, scaling and shifting parameters for each content feature point is aggregated from $V_{\sigma}$ and $V_{\mu}$ respectively according to the cross-attention map.
The process in one Transformer decoder block can be written as:
\begin{equation}
    \begin{aligned}
    F'_{cs}&=F_{cs}+{\rm MHA}(F_{cs},F_{cs},F_{cs}),\\
    \sigma&={\rm MHA}({\rm IN}(F'_{cs}),{\rm IN}(K_{s}),V_{\sigma}),\\
    \mu&={\rm MHA}({\rm IN}(F'_{cs}),{\rm IN}(K_{s}),V_{\mu}),\\
    F''_{cs}&=\sigma\odot F'_{cs}+\mu,\\
    F_{cs}&=F''_{cs}+{\rm MLP}(F''_{cs}),\label{eq:decoder}
    \end{aligned}
\end{equation}
where ${\rm IN}$ denotes instance normalization~\cite{ulyanov2016instance} and $\odot$ represents element-wise multiplication. 
$F_{cs}$ is initialized as $F_c$ before the first Transformer decoder layer. 

\subsection{Training Pipeline}
To achieve high-quality style transfer, 
we introduce a two-stage training strategy that comprises 
meta training and fast adaptation.
The meta training stage is designed to learn a generic model initialization,
while the fast adaptation adapts the network for a
single style
in a few iterations for few-shot style transfer. 
Note that zero-shot style transfer is a special case of the overall training configuration. 
\noindent\textbf{Meta Training:} Inspired by \emph{Reptile}~\cite{nichol2018firstorder}, a first-order meta learning algorithm, rendering style patterns of a specific reference style image can be viewed as a task. 
We seek an optimal initialization for neural networks in this stage, so that the networks can be rapidly adapted for a new task in only a few shots. 
The main training procedure for this stage is shown in Algorithm~\ref{reptile}. 
In each iteration, we sample $1$ style and $k$ batches of
contents to perform inner optimization to obtain ``fast weights'' $\omega$, which  
would later guide the update of ``slow weights'' $\theta$, to move a step in the direction of $\omega-\theta$.
Notably, as there is only one group of parameters for Transformer encoder and decoder layers, we randomly choose a number as the number of stacked layers for the Style Transformer in each iteration. 

\begin{algorithm}[!t]
  \caption{Meta Training}
  \label{reptile}
  \begin{algorithmic}[1]
    \Require
      $\mathcal{D}_c$: content dataset;
      $\mathcal{D}_s$: style dataset;
      $\delta$: inner learning rate;
      $\eta$: outer learning rate;
      $k$: number of inner updates;
      $T$: maximal number of stacked layers;
    \Ensure
      trained meta generator parameters $\theta$
    \State initialize $\theta$ randomly
     \For{iteration 1, 2, 3, $\cdots$}
      \State sample a style image $I_s$ from $\mathcal{D}_s$
      \State $\omega\leftarrow\theta$
      \For{$k$ times}
      \State sample a batch of content image $I_c$ from $\mathcal{D}_c$
      \State sample the number of layers from $1\sim T$
      \State forward propagation using Eq. \ref{eq_enc}-\ref{eq:decoder}
      \State compute inner loss $L$ using Eq. \ref{eq_loss}
      \State $\omega\leftarrow\omega-\delta \nabla L$
      \EndFor
      \State $\theta\leftarrow\theta+\eta(\omega-\theta)$
    \EndFor
  \end{algorithmic}
  
\end{algorithm}

\noindent\textbf{Fast Adaptation:} The trained model after the first stage would serve as an initialization and will be adapted for a single style. 
With the same objective, this stage behaves almost identically to the internal loop in Algorithm \ref{reptile}. 
The only difference is that only parameters of the Transformer encoder layer are necessary to be updated, since (1) the Transformer encoder layer is the main component to extract style patterns 
and has the most significant impact on stylization,
and (2) it would save memory and speed up the adaptation. 

\noindent\textbf{Zero-Shot Style Transfer:}
``Zero-Shot'' means that, once the meta training is done, there is no fast adaptation stage needed and 
the meta model itself can support arbitrary style transfer. 
To encourage the model to produce satisfactory results in the zero shot setup, we set the inner optimization time $k$ to 1, where the algorithm is reduced to the typical training paradigm of existing arbitrary style transfer methods. 
In this sense, Algorithm~\ref{reptile} provides a more general setting for style transfer in both zero-shot and few-shot cases.

\noindent\textbf{Text-Guided Style Transfer.} 
We perform text-guided style transfer based on our image style transfer model with slight modifications. 
Following the common practice, a pre-trained CLIP encoder is imported to extract text features. To be consistent with image input, we use a StyleGAN-like mapping network to convert text features into pseudo image features. In the meta training stage, since CLIP unifies the feature space, we use image instead of text as style input to avoid additional text dataset. In the fast adaptation stage, the mapping network along with the Transformer encoder is updated, considering there is still a gap between CLIP image feature and text feature.

\subsection{Loss Function}
The training objective in both meta-training and fast adaptation stages follows many works in arbitrary style transfer, which consists of content loss and style loss. 
Let $F^x$ represent features on \texttt{ReLU-x\_1} layer of a pre-trained VGG19 network~\cite{simonyan2015very} for loss computation. 
The content loss is defined by the normalized perceptual loss~\cite{johnson2016perceptual}:
\vspace{-0.3cm}
\begin{equation}
    \mathcal{L}_{cont}=\sum_{x=2}^5\Vert {\rm IN}(F_c^x)-{\rm IN}(F_{cs}^x)\Vert_2,\label{eq_cont}
\end{equation}
\vspace{-0.2cm}
while the style loss adopts the mean-variance loss~\cite{huang2017arbitrary}:
\begin{equation}
    \mathcal{L}_{sty}=\sum_{x=2}^5(\Vert \mu(F_s^x)-\mu (F_{cs}^x)\Vert_2 + \Vert \sigma(F_s^x)-\sigma(F_{cs}^x)\Vert_2),\label{eq_sty}
\end{equation}
where $\mu$ and $\sigma$ calculate the channel-wise mean and standard deviation separately.
The overall objective is given by the weighted summation of the two losses:
\vspace{-0.2cm}
\begin{equation}
    \mathcal{L}=\mathcal{L}_{cont}+\lambda\mathcal{L}_{sty},\label{eq_loss}
    \vspace{-0.2cm}
\end{equation}
where $\lambda$ controls the balance between two terms. 

In text-guided style transfer, the loss functions are the same as CLIPstyler, including global CLIP loss, directional CLIP loss and PatchCLIP loss. 

\section{Experiments}
\subsection{Implementation Details}
We use MS-COCO~\cite{lin2014microsoft} 
as our content dataset and WikiArt test set~\cite{phillips2011wiki} as our style dataset. 
The content dataset contains roughly 80,000 images and the style dataset has about 20,000 images. 
The optimizer is Adam~\cite{kingma2014adam} with learning rates of both inner and outer loops set as 0.0001 and the batch size is 4. 
In training, we first resize the content and style image to $512\times512$ and then randomly crop to $256\times256$ resolution. 
During inference, our model can handle inputs of any size.
The update times for inner optimization $k$ is set as 2 for few-shot case and 1 for zero-shot case.
In training, the maximal number of stacked layers is 4. 
All the multi-head attention blocks are instantiated as shifted window attention in \cite{liu2021swin}, with window size 8 and shift size 4.  
The model is trained on a Nvidia 3090 with 9k iterations in the meta training stage for convergence while only 100 steps for image input and 20 steps for text input in the fast adaptation for few-shot style transfer, which takes less than 1 minute. 
Hyper-parameter $\lambda$ is set as $10$. 

\begin{table}[!t]
\centering
    \begin{tabular}{cccc}
    \toprule
    Method & $\mathcal{L}_{cont}$ & $\mathcal{L}_{sim}$ & $\mathcal{L}_{sty}$ \\
    \midrule
        \multicolumn{1}{c}{AdaIN} & $4.88$\down{$\pm0.70$} & $0.53$\down{$\pm0.23$} & $1.51$\down{$\pm0.78$} \\
        \multicolumn{1}{c}{Linear} & $3.93$\down{$\pm0.76$} & $0.44$\down{$\pm0.17$} & $1.97$\down{$\pm0.96$} \\
        \multicolumn{1}{c}{AvatarNet} & $5.76$\down{$\pm0.53$} & $0.56$\down{$\pm0.19$} & $3.19$\down{$\pm1.82$} \\
        \multicolumn{1}{c}{SANet} & $4.72$\down{$\pm0.61$} & $0.50$\down{$\pm0.17$} & $1.15$\down{$\pm0.58$} \\
        \multicolumn{1}{c}{MANet} & $4.93$\down{$\pm0.57$} & $0.50$\down{$\pm0.17$} & $1.41$\down{$\pm0.77$}  \\
        \multicolumn{1}{c}{MCCNet} & $4.22$\down{$\pm0.69$} & $0.47$\down{$\pm0.17$} & $1.56$\down{$\pm0.84$} \\
        \multicolumn{1}{c}{AdaAttN} & $4.46$\down{$\pm0.70$} & $0.43$\down{$\pm0.16$} & $2.20$\down{$\pm1.19$} \\
        \multicolumn{1}{c}{StyTr2} & $3.78$\down{$\pm0.99$} & $0.48$\down{$\pm0.22$} & $1.50$\down{$\pm0.69$} \\
        \multicolumn{1}{c}{StyFormer} & $4.94$\down{$\pm0.79$} & $0.43$\down{$\pm0.15$} & $2.20$\down{$\pm1.44$} \\
        \multicolumn{1}{c}{MetaNet} & $\mathbf{3.48}$\down{$\pm0.85$} & $0.45$\down{$\pm0.19$} & $2.47$\down{$\pm1.69$} \\
        \multicolumn{1}{c}{MetaStyle$^*$} & $3.64$\down{$\pm1.12$} & $0.42$\down{$\pm0.18$} & $2.47$\down{$\pm1.06$} \\
        \multicolumn{1}{c}{Johnson$^\dag$} & $4.60$\down{$\pm0.76$} & $0.59$\down{$\pm0.20$} & $1.02$\down{$\pm0.34$} \\
        \hline
        \multicolumn{1}{c}{Ours-Vanilla} & $5.50$\down{$\pm0.63$} & $0.59$\down{$\pm0.26$} & $0.85$\down{$\pm0.38$} \\
        \multicolumn{1}{c}{Ours-Norm} & $4.70$\down{$\pm0.75$} & $0.43$\down{$\pm0.13$} & $0.93$\down{$\pm0.33$} \\
        \hline
        \multicolumn{1}{c}{Ours-ZS-L1} & $4.13$\down{$\pm0.68$} & $0.41$\down{$\pm0.14$} & $0.92$\down{$\pm0.40$} \\
        \multicolumn{1}{c}{Ours-ZS-L3} & $4.20$\down{$\pm0.68$} & $0.41$\down{$\pm0.13$} & $0.81$\down{$\pm0.31$} \\
        \multicolumn{1}{c}{Ours-FS$^*$} & $4.24$\down{$\pm0.82$} & $\mathbf{0.38}$\down{$\pm0.12$} & $\mathbf{0.79}$\down{$\pm0.25$} \\
    \bottomrule
    \end{tabular}
    \vspace{-0.3cm}
    \caption{Quantitative comparisons. \textit{ZS} and \textit{FS} for our model denote zero-shot and few-shot modes. \textit{Vanilla} denote replacing our architecture with original Transformer. \textit{Norm} means adding layer normalization in the Transformer encoder layer. \textit{L1}/\textit{L3} means using 1/3 Transformer layers in the test time. $*$ and $\dag$ denote few-shot and per-style-per-model methods.}
    \vspace{-0.5cm}
    \label{table:loss}
\end{table}

\subsection{Comparison with Prior Works}
In this section, we compare results by our Master with 13 state-of-the-art style transfer methods, including 3 global transformation based methods (AdaIN~\cite{huang2017arbitrary}, Linear style transfer~\cite{li2018learning}, and MCCNet~\cite{deng2020arbitrary2}), 1 patch swap based method (Avatar-Net~\cite{sheng2018avatar}), 3 attention based methods (SANet~\cite{park2019arbitrary}, MANet~\cite{deng2020arbitrary1}, and AdaAttN~\cite{Liu_2021_ICCV}), 2 transformer based methods (StyTr2~\cite{deng2021stytr2} and StyFromer~\cite{wu2021styleformer}), 2 meta learning based methods (MetaNet~\cite{shen2017meta} and MetaStyle~\cite{zhang2019metastyle}), 1 per-style-per-model method by Johnson \textit{et al.}~\cite{johnson2016perceptual} and 1 text-guided style transfer method (CLIPstyler). 

\noindent\textbf{Quantitative Comparison.}
We adopt content loss $\mathcal{L}_{cont}$ in Eq. \ref{eq_cont} and style loss $\mathcal{L}_{sty}$ in Eq. \ref{eq_sty} as evaluation metrics to reflect effects of learning by different methods. 
We also design a metric $\mathcal{L}_{sim}$ to reflect the preservation of the spatial-wise self cosine similarity of content structures:
\vspace{-0.3cm}
\begin{equation}
\begin{aligned}
    D_{*,ij}^x=1&-\frac{F_{*,i}^x\cdot F_{*,j}^x}{\Vert F_{*,i}^x\Vert \Vert F_{*,j}^x\Vert}\\
    \mathcal{L}_{sim}=\sum_{x=3}^4\frac{1}{n_x^2}\sum_{i,j}&\left| \frac{D_{c,ij}^x}{\sum_{k}D_{c,kj}^x}-\frac{D_{cs,ij}^x}{\sum_{k}D_{cs,kj}^x} \right|,
\end{aligned}
\vspace{-0.2cm}
\end{equation}
where $n_x$ is the number of spatial locations for the current feature map and the second foot script denotes the spatial index. 
Smaller $\mathcal{L}_{sim}$ means that the original spatial-wise relationship is better preserved during style transfer, \textit{i.e.}, less content distortion. 
We use the test dataset in the code page of \cite{huang2017arbitrary}
for evaluation, with 11 content images and 20 style images to form 220 content-style pairs. 
We report the mean and standard deviation over the 220 cases in Tab. \ref{table:loss}. 
Notably, the lowest style loss and content similarity loss are achieved by Master simultaneously compared with previous methods, even better than the per-style-per-model solution by Johnson \textit{et al.}~\cite{johnson2016perceptual}, and can be further reduced in the few-shot case with comparable content loss values, which demonstrates the \textbf{joint} advantages of our method in content preserving and style rendering. 
Meanwhile, the lower standard deviations demonstrate the robustness of our model. 

\begin{figure}
  \includegraphics[width=\linewidth]{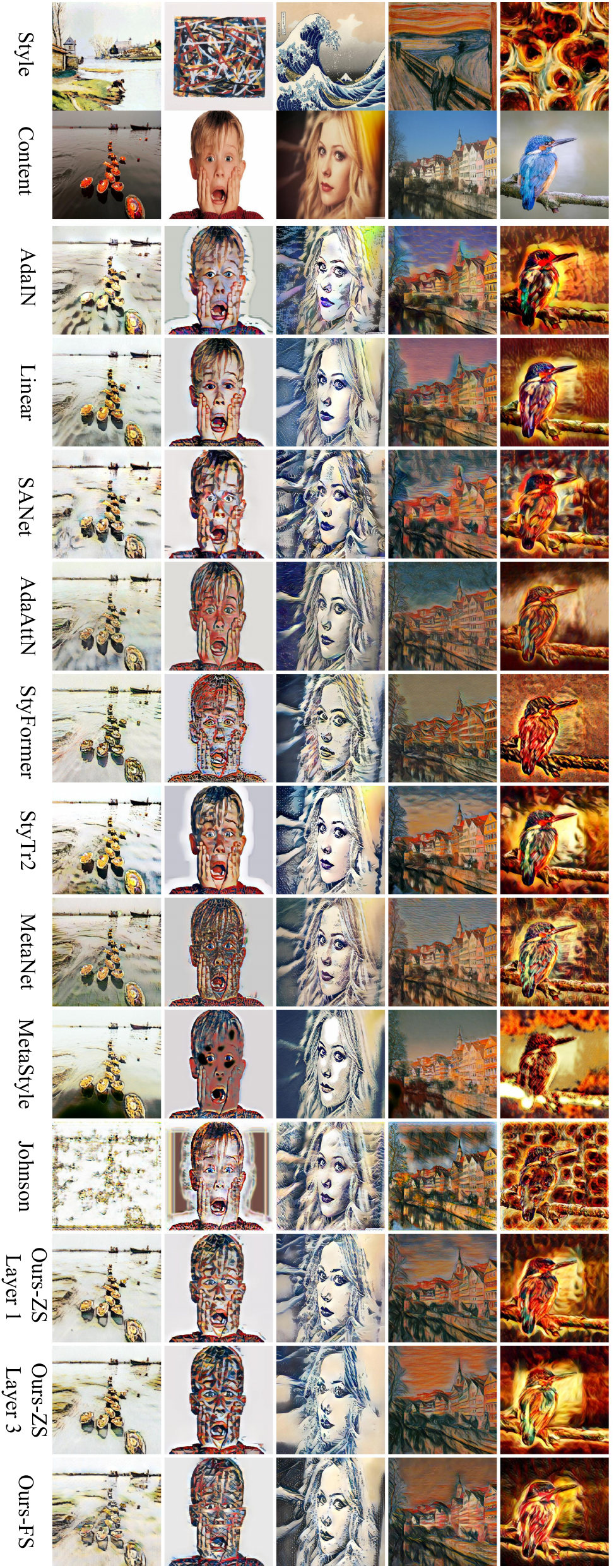}
  \caption{Comparison with previous state-of-the-art style transfer methods. Zoom in for better details.}
  \label{fig:compare_sota}
\end{figure}

\begin{figure}
  \includegraphics[width=\linewidth]{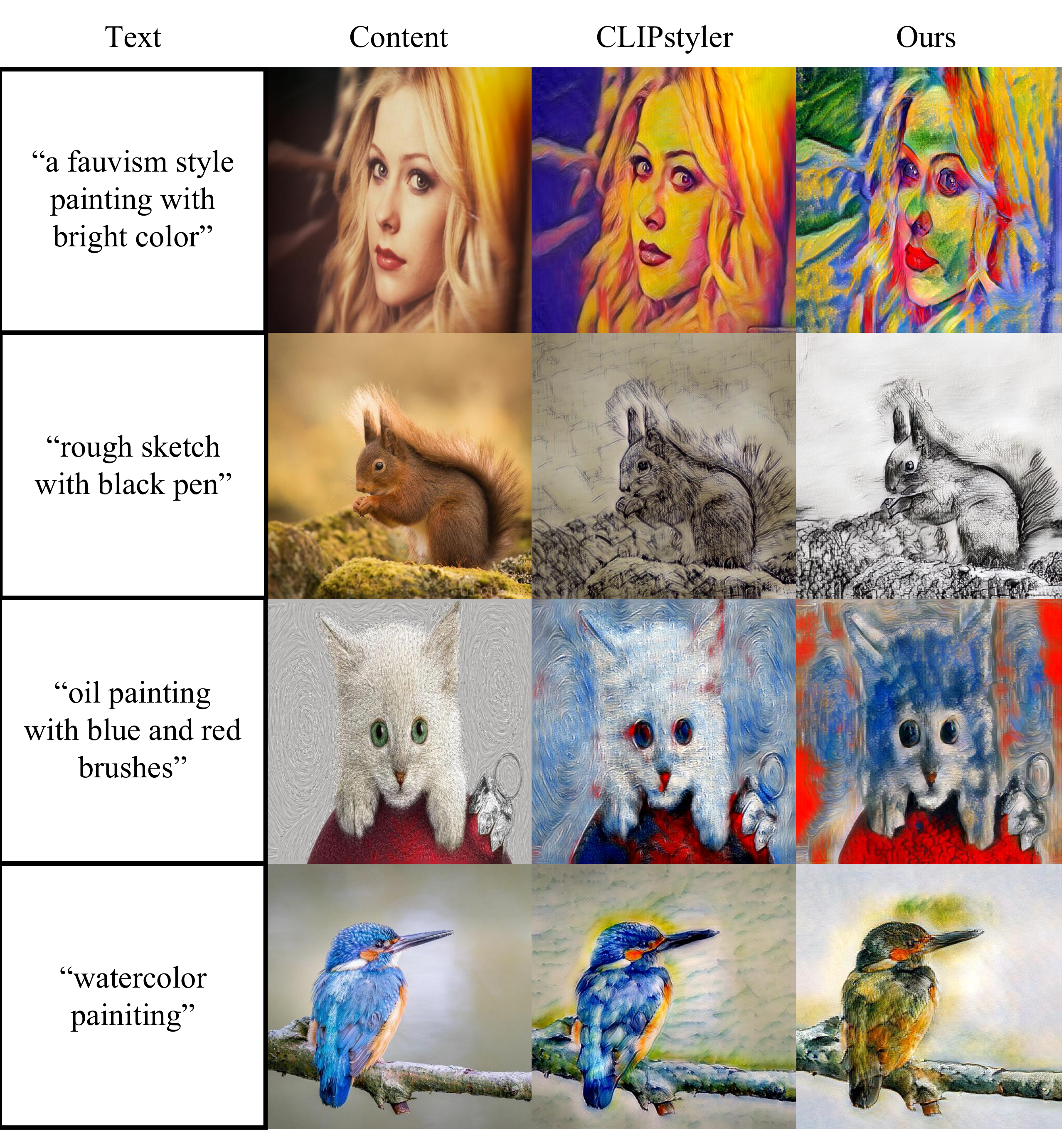}
  \caption{Comparison with CLIPstyler in text-guided style transfer. Zoom in for better details.}
  \vspace{-0.7cm}
  \label{fig:compare_text}
\end{figure}

\noindent\textbf{Qualitative Comparison.}
Qualitative examples by different methods are shown in Fig. \ref{fig:compare_sota}. 
Full comparisons with more methods can be found in the supplementary material. 

We first give discussion on the global-transformation based methods. 
As shown in the 3rd and 4th rows, in AdaIN and Linear, features in all positions share the same transformation matrix, which fails to migrate detail textures in style images and has poor color saturation. 

As for the attention-based methods, SANet in the 5th row introduces local focus to style images. 
However, due to the simple fusion of one-layer attention results and original content features, it brings textures and content distortion in many cases, \textit{e.g.}, background areas of the 4th and 5th columns. 
AdaAttN in the 6th row dedicates to addressing the distortion problem in SANet, but it sacrifices the ability to render salient style patterns and textures. 

We then compare our results with those by Transformer-based methods. 
StyFormer in the 7th row adopts a design of the cross-attention module in Transformer without exploring the effect of self-attention to enhance style rendering. 
Thus, the results appear under-stylized compared with those by our method. 
StyTr2 in the 8th row is a Transformer architecture with a pure self-attention mechanism, which makes the performance on the extraction and migration of local textures not satisfactory enough, \textit{e.g.}, strokes of the 2nd column and waves of the 4th column. 
Moreover, since it follows the design of residual connections in vanilla Transformer, the issue of content distortion in Fig. \ref{fig:motivation} is still evident, as shown in the background areas of the 4th and 5th columns. 
Compared with these methods, Master renders more vivid global and local style patterns. 
Meanwhile, it well maintains content structures simultaneously with learnable scaling and shifting parameters. 
Furthermore, in the few-shot case, comparable performance can be achieved with only 1 Transformer layer. 

MetaNet and MetaStyle also involve the concept of meta learning into their style transfer frameworks. 
On the one hand, MetaNet predicts parameters of a generator network given one style image via a meta network, which is an ambitious goal and makes the training more difficult. 
Thus, as shown in the 9th row, the stylized effects of their results still appear weak. 
On the other hand, MetaStyle achieves inferior performance in terms of local details as demonstrated in the 10th row, since only parameters of global normalization layers are learnable in the fast adaptation stage. 

The single style transfer method by Johnson \textit{et al.} in the 11th row tends to arrange patterns learned by the network arbitrarily, which may produce undesired and distorted effects. 
By contrast, the complex content-style dependence is constructed in our Transformer model and thus achieves more remarkable performance on transferring style patterns to proper positions in content images. 

Some qualitative comparisons on text-guided style transfer with CLIPstyler are shown in Fig. \ref{fig:compare_text}, which uses per-text-per-model fashion. 
Through these results, we observe that despite significantly more training steps, CLIPstyler still suffers from weak stylization. 
By contrast, our method generates more vivid results in only 20 adaptation steps, which demonstrates the advantages of the proposed Master architecture and the versatility of our training pipeline. 

\begin{figure}[!t]
  \centering
  \includegraphics[width=0.8\linewidth]{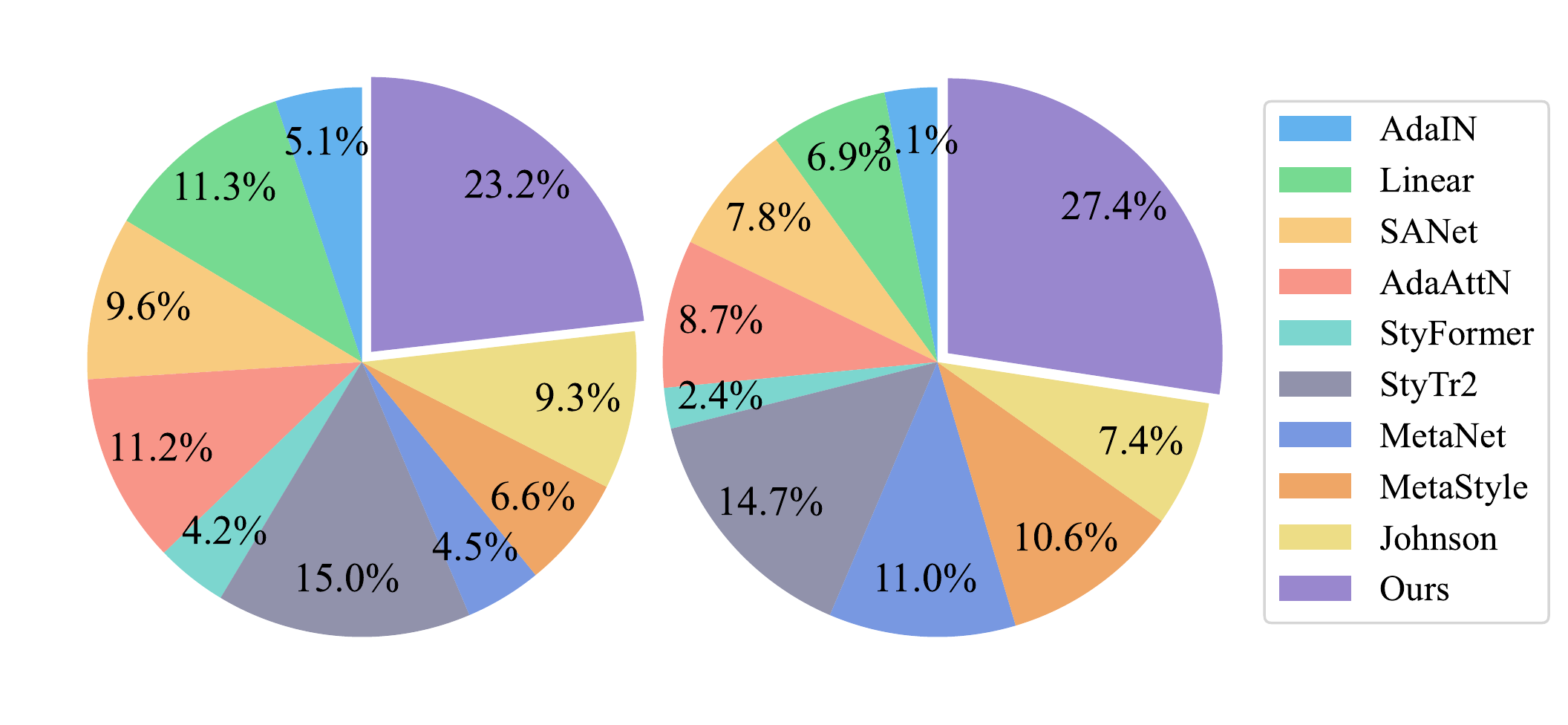}
  \vspace{-0.2cm}
  \caption{Results of user study. Left: Our zero-shot stylization results compared with SOTA methods. Right: Our few-shot  stylization comparison results.}
  \vspace{-0.3cm}
  \label{fig:user_study}
\end{figure}

\noindent\textbf{User Study.}
Following most style transfer works, we conduct user study and report user preference. 
We choose 20 content images and 20 style images to form 400 content-style pairs.
We involve 100 people and
randomly assign 20 stylized results from compared methods to each subject.
Our method showcases zero-shot results with 3 Transformer layers in 10 tests and few-shot results in the remaining 10.
For each pair, the order of results is randomized, and participants choose their favorite. 
With 2,000 votes in total, Fig. \ref{fig:user_study} demonstrates Master's superior style transfer quality.

\begin{figure}
\centering
  \includegraphics[width=\linewidth]{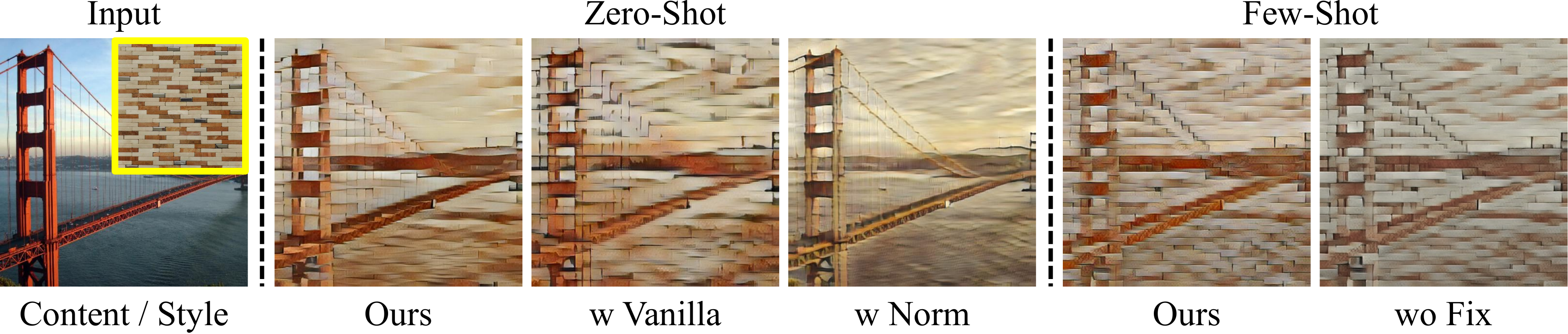}
  \vspace{-0.6cm}
  \caption{Ablation study on key designs in Master model.}
  \vspace{-0.6cm}
  \label{fig:ablation}
\end{figure}

\subsection{Ablation Study}

\textbf{Architecture.}
We conduct ablation studies on key designs in Master to illustrate their effects on the stylization quality. 
As shown in Fig. \ref{fig:ablation}, on the one hand, if vanilla Transformer model is used, without learnable scaling and shifting parameters, noisy textures can be introduced significantly, which distorts the original content structures and affects style transfer quality. 
On the other hand, if layer normalization operations in the standard Transformer are used in the style encoder, style patterns would become less saturated, since normalization removes second-order feature statistics which represents the style information to a large degree. 
Quantitative studies in Tab. \ref{table:loss} also support our analysis and  come to the same conclusion.

\begin{figure}[!t]
    \centering
    \includegraphics[width=0.8\linewidth]{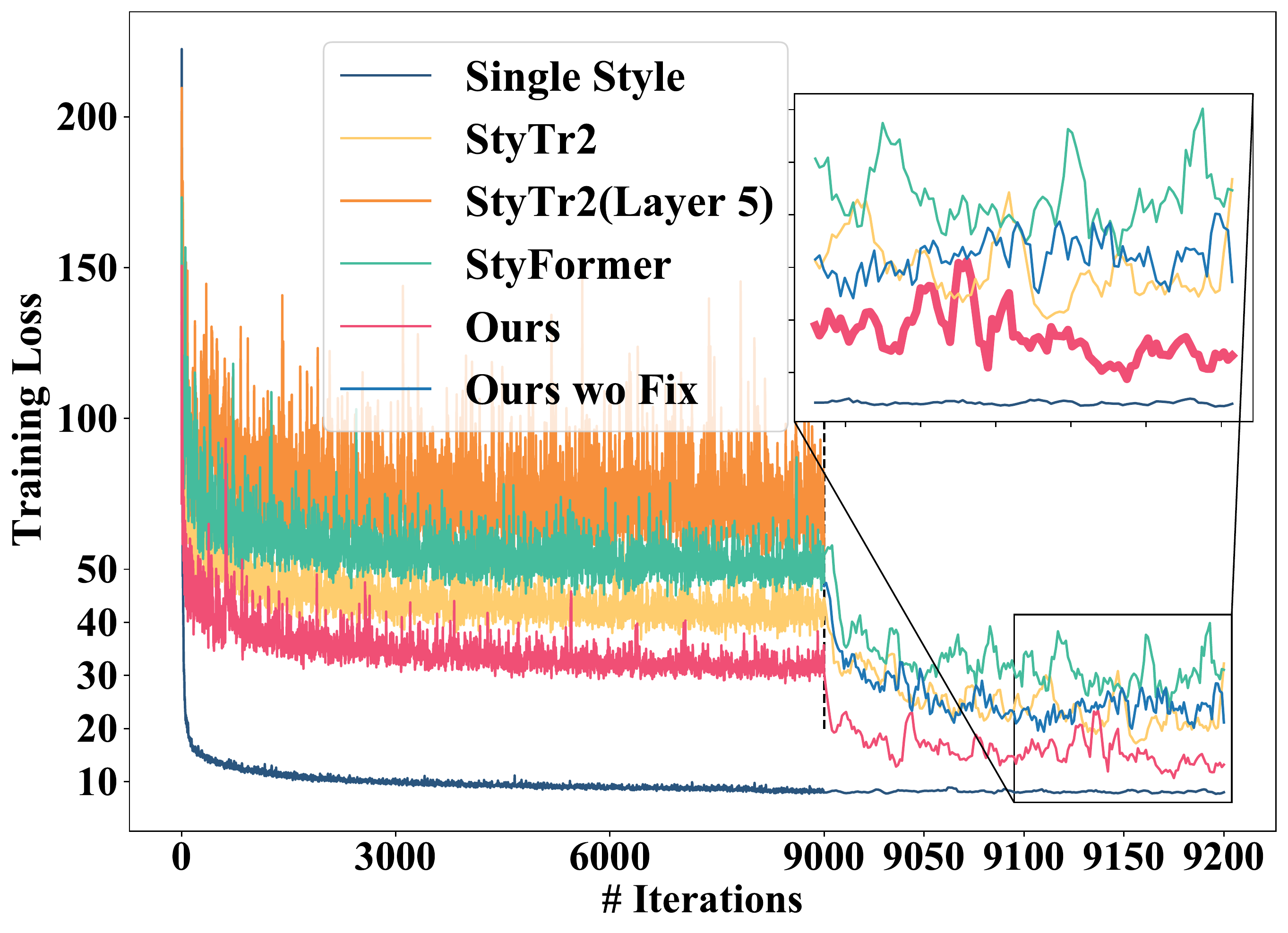}
    \vspace{-0.3cm}
    \caption{Training loss visualization during meta training and fast adaptation of three transformer based models: StyTr2, StyFormer, and our Master. The loss curve under per-style-per-model setting by our model is used as a reference. \textit{Ours-wo-Fix} means that all the parameters need to be updated during the fast adaptation stage.}
    \vspace{-0.6cm}
    \label{fig:fast_adapt}
\end{figure}

\noindent\textbf{Meta Training and Fast Adaptation.} 
We provide training loss visualization in Fig. \ref{fig:fast_adapt}, to demonstrate the effectiveness of our meta training and fast adaptation algorithm. 
Our meta training stage learns an appropriate initial state, which enables fast decent and convergence of loss values in the fast adaptation stage. 
We show the same visualization in the per-style-per-model setting as a benchmark, using the same loss function and network architecture. 
As shown in Fig. \ref{fig:fast_adapt}, it requires roughly 3k iterations for the model to be convergent in the per-style-per-model case (Single Style), while the meta model (Ours) after 9k iterations can be adapted for any style image in only a few shots with competitive training results.
Thus, the total number of iterations for our method is significantly smaller than per-style-per-model ones with the growth number of required styles. 

By default, all the parameters except those in the style encoder module are fixed during the fast adaptation stage. 
We also experiment with training the whole model in this stage (Ours wo Fix in Fig. \ref{fig:fast_adapt}) and find that the training and convergence would become difficult and require more time and memory. 
Moreover, updating all the parameters may result in insufficient adaptation given a limited number of training steps.
As demonstrated in Fig. \ref{fig:ablation}, there is a gap on the global tone between the result and the style image.

\noindent\textbf{Base Model.} 
We evaluate StyFormer and StyTr2 as base models, presenting loss visualizations. Heavier Transformers raise training difficulty for earlier models, resulting in inferior convergence.
When the number of Transformer layers is 5, StyTr2 even fails to converge. 
By contrast, our model adopts parameter-sharing across all Transformer layers, which results in an overall light-weight structure. 
Consequently, it enjoys better training effects in the meta training stage as shown in Fig. \ref{fig:fast_adapt}. 
Moreover, as shown in the zoom-in part, Master also enables overall lower and more stable training in fast adaptation. 
Qualitative examples by different base models can be found in the supplement. 

\begin{figure}[!t]
    \centering
    \includegraphics[width=0.9\linewidth]{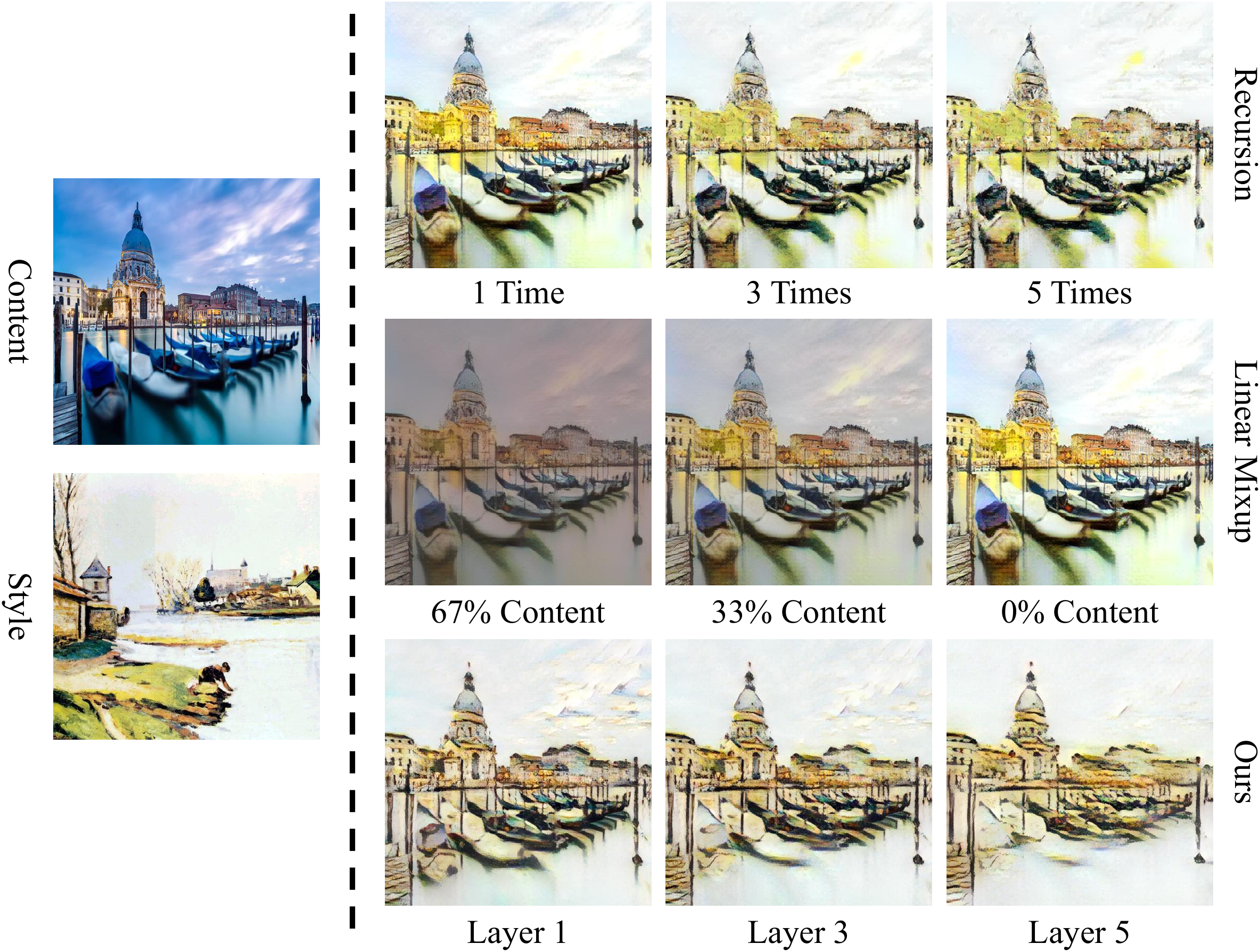}
    \vspace{-0.3cm}
    \caption{Comparisons with widely-adopted approaches to control the degree of stylization.}
    \vspace{-0.6cm}
    \label{fig:control}
\end{figure}

\noindent\textbf{Controllable Style Transfer.} 
We compare with some widely-adopted approaches for the similar goal to control the degree of stylization, including (1) Recursion, which treats stylized results as input of content images in the next iteration, and (2) Linear Mixup, which conducts linear combinations between content features and stylized features before the decoder. 
Vanilla Transformer is adopted for these approaches. 
The major difference between Recursion and ours is that our method takes the iterative stylization into consideration in the training time so that the model would learn to transfer style patterns layer-by-layer. 
Thus, as shown in Fig. \ref{fig:control}, Recursion merely uses more intensive colors with the increasing of repeat times. 
Also, for Linear Mixup, it is typically hard to generate reasonable style transfer results especially when the weight of contents is high. 
Compared with them, our method is capable of rendering more style patterns and increasing the artistic abstraction when more Transformer layers are stacked. 

\section{Conclusion}
In this paper, we propose a novel Transformer model specifically for artistic style transfer, termed as Meta Style Transformer (Master). 
On one hand, parameters of different Transformer layers are shared, which reduces the total number of parameters significantly and thus enables easier convergence. 
It is also convenient for Master to control the degree of stylization via customizing the number of stacked layers in the inference time. 
On the other hand, different from standard Transformer, our model adopts dynamic and learnable scaling and shifting operations instead of original residual connections, which helps preserve similarity relationship in content structures while migrating remarkable style patterns. 
Specifically, Master is trained using a meta learning algorithm for few-shot style transfer. 
Only parameters of the Transformer encoder layer need to be updated in the few-shot stage, which benefits the fast and robust adaptation. 
Zero-shot arbitrary style transfer is a special case of the training configuration. 
Experiments suggest that Master outperforms previous state-of-the-art arbitrary style transfer methods on both content preserving and style rendering. 
 
\section*{Acknowledgment}
This project is supported by the Singapore Ministry of Education Academic Research Fund Tier~1 (WBS: A-0009440-01-00).

{\small
\bibliographystyle{ieee_fullname}
\bibliography{master}
}

\clearpage

\appendix
Here, we provide more experimental analysis and results of the proposed meta style transformer (Master) for controllable zero-shot and few-shot artistic style transfer, which cannot be accommodated in the main paper due to the page limitation. 
We first compare our model with the existing Transformer-based methods in terms of efficiency. 
Then, we provide some qualitative analysis and ablation studies to the meta training and fast adaptation algorithms. 
Finally, we supplement more comparisons with more state-of-the-art techniques, more zero-shot and few-shot style transfer results, more examples of controlling the stylization via stacking different numbers of Transformer layers, more results of text-guided style transfer, and more extensions. 

\begin{table}[!t]
\centering
    \begin{tabular}{ccc}
    \toprule
    Setting & Speed (sec. / image) & \# of Param. (M) \\
    \midrule
        StyTr2 & 0.087 & 25.14 \\
        \hline
        Ours-L1 & 0.024 & 10.75 \\
        Ours-L3 & 0.030 & 10.75 \\
        Ours-L5 & 0.038 & 10.75 \\
    \bottomrule
    \end{tabular}
    \caption{Comparisons on inference speed and number of parameters at different settings. StyTr2 adopts 3 Transformer layers by default. For our method, L1/L3/L5 means using 1/3/5 Transformer layers in the test time.}
    \label{table:efficiency}
\end{table}

\begin{table}[!t]
\centering
    \begin{tabular}{ccccc}
    \toprule
    $k$ & 1 & 2 & 3 & 4 \\
    \midrule
    $\mathcal{L}_{sty}$ & 3.389 & 2.661 & 2.384 & 1.811 \\
    \bottomrule
    \end{tabular}
    \caption{Impact of the number of inner optimization times $k$ in the meta training on the style loss in the fast adaptation.}
    \label{table:k}
\end{table}

\section{Efficiency}
In this part, we compare the proposed Master model with the state-of-the-art Transformer-based style transfer method StyTr2~\cite{deng2021stytr2}, in terms of inference speed and number of parameters. 
We experiment with stacking 1, 3, and 5 Transformer layers in the test time and comparisons at different settings are shown in Tab. \ref{table:efficiency}. 
Here, StyTr2 adopts 3 Transformer layers by default and comparisons are conducted under $512\times512$ resolution. 
The speed is measured over 220 inference times and the same workstation with a Nvidia 3090 GPU is adopted as the platform for all settings. 

Through the results, we can observe that the proposed model can have more than $2\times$ FPS compared with StyTr2, even when the number of Transformer layers is 5. 
Moreover, since parameters are shared across different Transformer layers, the total number of parameters would not increase with the increasing number of stacked layers and it is always significantly less than that of StyTr2. 
Thus, compared with existing Transformer-based models, Master achieves superior quality and efficiency simultaneously. 

\begin{figure}[t]
  \includegraphics[width=\linewidth]{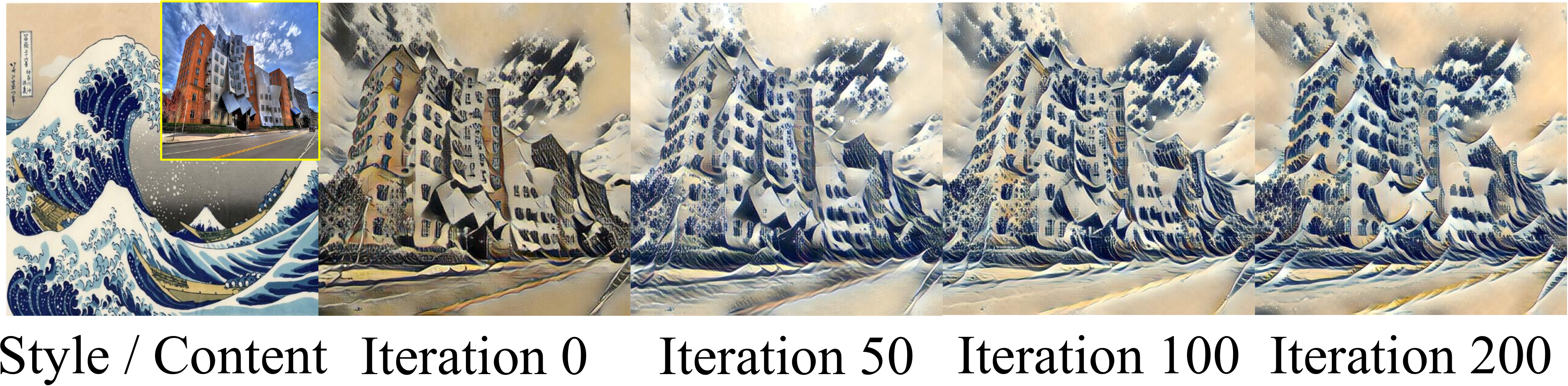}
  \caption{Impact of the number of fast adaptation iterations.}
  \label{fig:fast_adapt_example}
\end{figure}

\begin{figure}[t]
  \includegraphics[width=\linewidth]{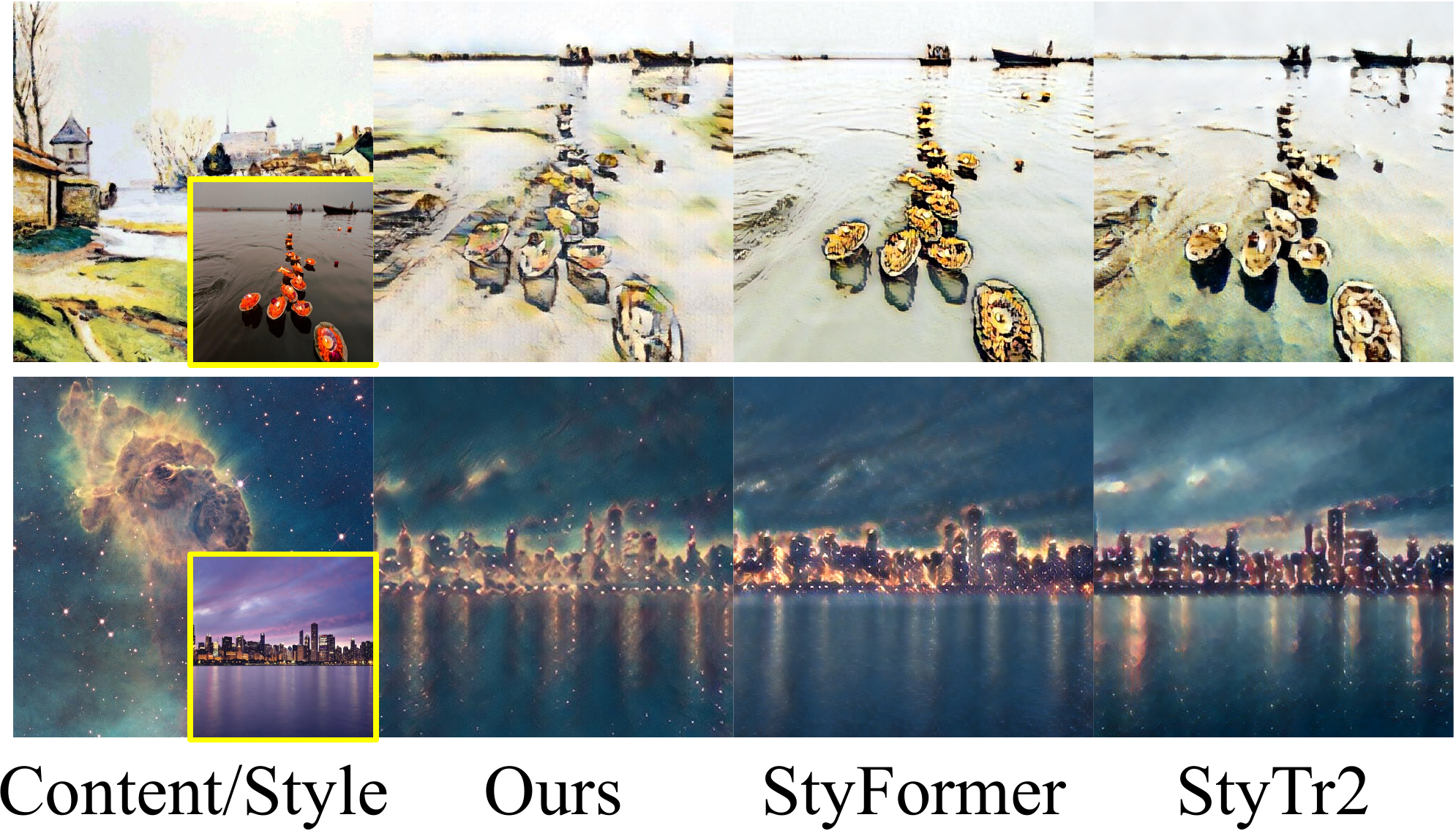}
  \caption{Few-shot stylization results under different base models: Our Master, StyFormer, and StyTr2.}
  \label{fig:baseline}
\end{figure}

\begin{figure}[t]
\centering
  \includegraphics[width=\linewidth]{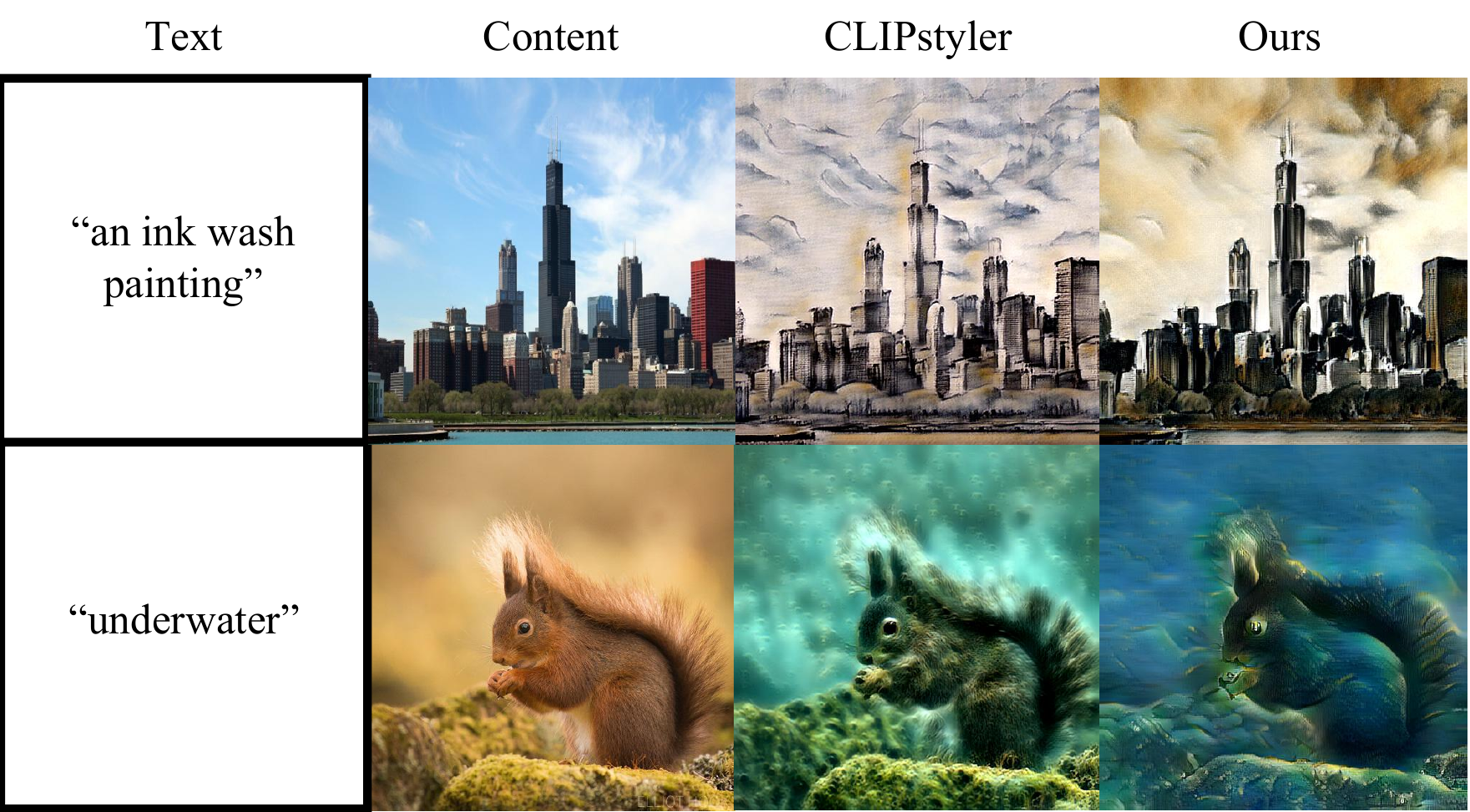}
  \caption{More comparisons on text-guided style transfer with Clipstyler.}
  \label{fig:compare_text_supp}
\end{figure}

\section{Meta Training and Fast Adaptation}

\noindent\textbf{Meta Training.} 
Alg. 1 of the main paper shows the workflow of the meta training procedure, and the number of inner optimization times $k$ is a hyper-parameter. 
As a meta learning algorithm, it requires $k$ to be greater than $1$. 
Otherwise, it would be degraded into pretraining and fine-tuning~\cite{nichol2018firstorder}, which is essentially equivalent to the typical training pipeline in arbitrary style transfer and is the setting in our zero-shot style transfer. 
The larger the $k$ is, the higher order of optimization procedures in few-shot learning can be learned, which may contribute to faster adaptation in the few-shot stage. 
Results of Tab. \ref{table:k}, which show the average style loss at the 10th iteration of fast adaptation, validate this effect. 

\noindent\textbf{Fast Adaptation.} 
We provide example results of different numbers of fast adaptation iterations in Fig. \ref{fig:fast_adapt_example}.
More local and global style patterns are captured by our model with the progress of fast adaptation, which suggests that our method potentially supports {user-customized level} of stylization by controlling the number of iterations during fast adaptation. 
Specifically, in all our experiments, we adopt 100 as the default number of iterations during the fast adaptation stage.

To further demonstrate the advantage of the Transformer model, we change our base model from our architecture to StyFormer and StyTr2 respectively and provide qualitative examples by these base models in Fig. \ref{fig:baseline}, as a supplement to the training analysis in Fig. 7 of the main paper.
We observe that our model renders global and local style patterns better. 

\section{More Results}

\subsection{Full Comparison Results} 
In order to further demonstrate the advantages of our proposed method, we provide more comparisons between our results with those by more state-of-the-art methods, as a supplement to Fig. 4 in the main paper. 
Here, there are 3 global transformation based methods (AdaIN~\cite{huang2017arbitrary}, Linear style transfer~\cite{li2018learning}, and MCCNet~\cite{deng2020arbitrary2}), 1 patch swap based method (Avatar-Net~\cite{sheng2018avatar}), 3 attention based methods (SANet~\cite{park2019arbitrary}, MANet~\cite{deng2020arbitrary1}, and AdaAttN~\cite{Liu_2021_ICCV}), 2 transformer based methods (StyTr2~\cite{deng2021stytr2} and StyFromer~\cite{wu2021styleformer}), 2 meta learning based methods (MetaNet~\cite{shen2017meta} and MetaStyle~\cite{zhang2019metastyle}), and the per-style-per-model method by Johnson \textit{et al.}~\cite{johnson2016perceptual}. 
The comparisons are shown in Fig. \ref{fig:compare_supp} and the conclusions are consistent with those in the main paper:
\begin{itemize}
\item Global transformation based methods are not powerful enough to capture local style details. 
\item The patch based method Avatar-Net distorts major content structures heavily. 
\item Attention based methods are prone to either dirty textures, \textit{e.g.}, SANet and MANet, or shallow style pattern migration, \textit{e.g.}, AdaAttN. 
\item Following the design of vanilla Transformer, similar problems of dirty textures and content distortion also exist in StyTr2, \textit{e.g.}, 4th, 5th, 6th, and 10th columns. 
Moreover, without leveraging local transformation, its performance on migration of local textures is not satisfactory enough, \textit{e.g.}, 1st, 2nd, 3rd, 7th, 8th, 9th, and 11th columns. 
\item Compared with StyFormer, the local self-attention mechanism in our model extracts and transfers style patterns more sophisticatedly. 
\item It seems hard for MetaNet to be robustly adapted for a style image in a few shots. 
\item Results by MetaStyle often demonstrate shallower stylized effects compared with ours.
\item Johnson \textit{et al.} tends to fill content images with the learned style textures, which may also distort content structures. The similar effect also exists in the comparison results with the seminal optimization-based solution by Gatys \textit{et al.}~\cite{Gatys_2016_CVPR} as shown in Fig.~\ref{fig:more}(c) and Tab.~\ref{tab:more}. 
\end{itemize}
Our method addresses above problems by dedicated self attention and cross-modality attention mechanisms with learnable and dynamic scaling parameters, which lead to more robust and vivid stylization results.

\subsection{More Content-Style Pairs} 
To further illustrate the performance of our Master model, we provide more content-style pairs in Fig. \ref{fig:pair}. 
In each entry, upper and bottom images are results under zero-shot and few-shot settings respectively. 
Here, 1 Transformer layer is adopted. 
These results better demonstrate the robustness of our method to different kinds of content and style images. 

\subsection{More Controllable Style Transfer Results}
We provide more controllable style transfer results by using different numbers of stacked Transformer layers in the inference time. 
As shown in Fig. \ref{fig:control_supp}, with more Transformer layers executed, the degree of stylization increases in general, where more intensive and vivid global and local style patterns are migrated. 
Quantitatively, we visualize the effect of tuning the number of Transformer layers in Fig.~\ref{fig:hyper}, which demonstrates that the trade-off between content loss and style loss can be controlled by this factor. 

\subsection{More Text-Guided Style Transfer Results}
As a supplement to Fig. 6 in the main manuscript, in Fig. \ref{fig:compare_text_supp}, we provide more qualitative comparison with Clipstyler~\cite{kwon2022clipstyler}, the state-of-the-art text-guided style transfer technique based on the per-text-per-model fashion. 
The conclusion is consistent with that in the main paper. 
We also visualize more pair-wise results of different texts and content images in Fig. \ref{fig:text_pair}. 

\subsection{More Ablation Results} 

\textbf{Architecture:} 
We provide more ablation results to better support the necessity of key designs in our Master model: using learnable scaling parameters for cross-attention, removing normalization in style encoder, and only updating style encoder in the few-shot training stage. 
The results are shown in Fig. \ref{fig:ablation_supp}, as a supplement to Fig. 6 in the main paper. 
Through the results, we can observe:
\begin{itemize}
\item Vanilla Transformer without learnable scaling parameters tend to distort original content structures. Such effects are obvious in background areas with less variation on textures. 
\item Using normalization in style encoder is harmful for stylization effects, since second-order statistics removed by the normalization contain important style information. 
\item Updating the whole model in the few-shot stage makes the training more difficult and leads to inferior stylization effects, compared with the case of only updating style encoder. 
\end{itemize}

\begin{figure}[t]
\centering
  \includegraphics[width=\linewidth]{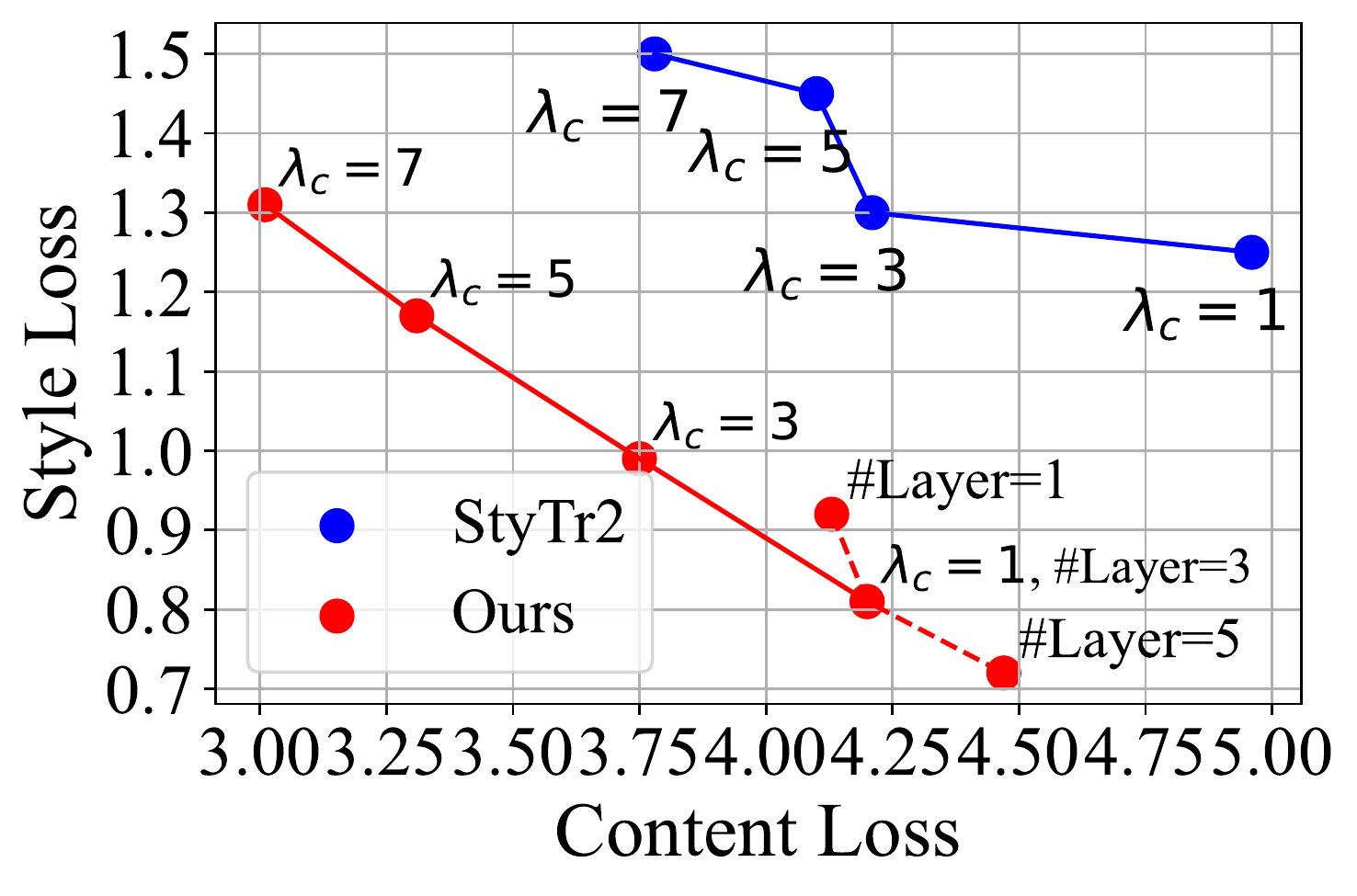}
  \caption{Quantitative comparisons with StyTr2 under different configurations of content weight.}
  \label{fig:hyper}
\end{figure}

\textbf{Hyper-Parameter:} 
For a fair comparison, we compare with StyTr2~\cite{deng2021stytr2}, the vanilla Transformer model for style transfer, under the same configuration of loss function, \textit{i.e.}, the same content weight, denoted as $\lambda_{c}$. 
The default $\lambda_{c}$ in this paper is $1$ while that in StyTr2 is $7$, and the quantitative results are shown in Tab.1 of the main paper. 
The results under the same $\lambda_{c}$ are provided in Fig.~\ref{fig:hyper}, where the superiority of our method can be reflected more clearly. 

\begin{figure}[t]
\centering
  \includegraphics[width=\linewidth]{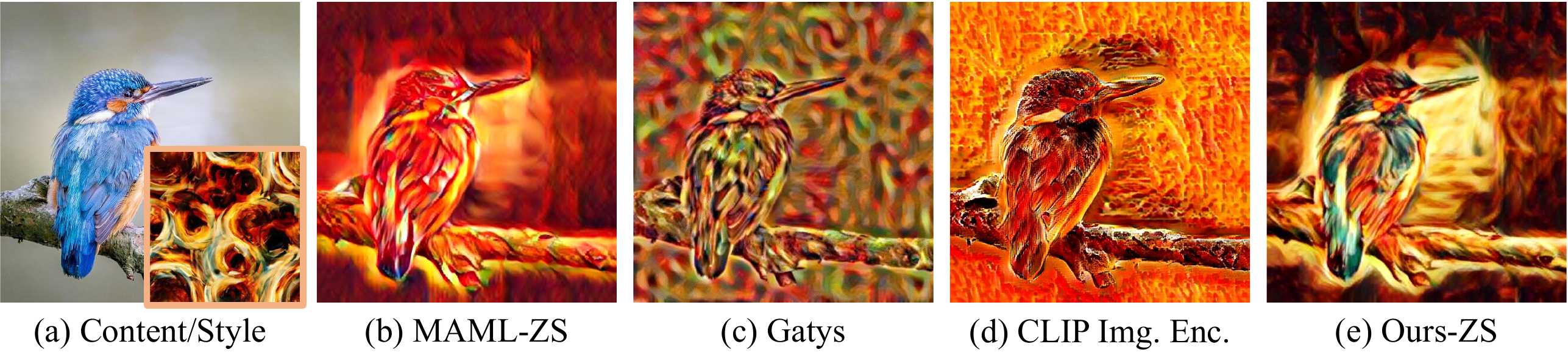}
  \caption{More qualitative ablation studies.}
  \label{fig:more}
\end{figure}

\textbf{Training Algorithm:} 
Compared with MetaStyle~\cite{zhang2019metastyle}, a MAML-based few-shot style transfer method, our method has two major differences: the training algorithm is based on Reptile and the architecture is a novel Transformer model. 
We provide a fine-grained ablation study in Fig.~\ref{fig:more} and Tab.~\ref{tab:more}, both qualitatively and quantitatively, to reflect the contribution of each component. 
In fact, both the model and the algorithm make improvement: the Transformer model mainly improves the stylization quality compared with existing models while Reptile mainly improves the training efficiency compared with MAML in MetaStyle. 
On the one hand, as shown in Fig.~8 of the main paper, replacing Master with vanilla Transformers would result in inferior quantitative metrics. 
On the other hand, we tried using MAML instead of Reptile before and found that it requires more time for convergence: 3 days for MAML v.s. 5 hours for Reptile. 
The computation of higher-order gradients increases the training difficulty, which further results in inferior performance as shown in Fig.~\ref{fig:more}(b) and Tab.~\ref{tab:more}. 
We also include ArtFID~\cite{wright2022artfid}, a recently proposed metric for artistic style transfer, for better illustration. 

\textbf{Encoder:}
Our method adopts CLIP~\cite{radford2021learning} to achieve text-guided style transfer, which contains an image encoder and text encoder. 
We use the image encoder for training and adopt the text encoder for inference, leveraging the aligned feature spaces of corresponding images and texts. 
In fact, it is also feasible to use the CLIP image encoder for image style transfer, rather than the Swin encoder by default. 
An example is shown in Fig.~\ref{fig:more}(d). 
Since CLIP only returns a 512-d feature vector for an image, it mainly transfers the style globally and the performance on local details is inferior. 
Thus, Swin is used for image style transfer by default. 

\begin{table}[!t]
\centering
    \begin{tabular}{ccccc}
    \toprule
    & & $\mathcal{L}_{cont}\downarrow$ & $\mathcal{L}_{sty}\downarrow$ & ArtFID$\downarrow$ \\
    \midrule
    \multicolumn{2}{c}{Gatys \textit{et al.}} & 4.24 & 1.67 & 37.24 \\
    \multicolumn{2}{c}{StyTr2 (Same $\lambda_c$)} & 4.96 & 1.25 & 40.49 \\
    \multirow{2}{*}{\makecell[c]{MAML}} & ZS & 4.95 & 2.36 & 38.14 \\
    & FS & 4.80 & 0.79 & 34.47 \\
    \multirow{2}{*}{\makecell[c]{Ours}} & ZS & \textbf{4.20} & \textbf{0.81} & \textbf{32.80} \\
    & FS & \textbf{4.24} & \textbf{0.79} & \textbf{32.70} \\
    \bottomrule
    \end{tabular}
    \caption{More quantitative ablation studies.}
    \label{tab:more}
\end{table}

\textbf{Content-Distortion Problem:} 
We provide a more specific example to illustrate the content-distortion problem by the vanilla Transformer model. 
Assume that there are two 2-d content features: $c_1=[0.5,1]$ and $c_2=[4,1.5]$, two style features: $s_1=[3.5,0]$ and $s_2=[-5,-5]$. 
Attention scores after Softmax are close to $1$ for both $c_1$ and $c_2$ to $s_1$, and are close to $0$ for both $c_1$ and $c_2$ to $s_2$. 
The transferred results with residual connection are $cs_1=[4,1]$ and $cs_2=[7.5,1.5]$, and the cosine similarity between $c_1$ and $c_2$ becomes $1$ from $0.73$. 
Thus, the original content-wise similarity is distorted. 
In this case, re-scaling content features by a factor larger than $1$ may alleviate the drawback. 
This factor is made learnable in this paper and the model is provided with an opportunity to learn how to preserve the similarity in training and convergence. 
The metric $\mathcal{L}_{sim}$ in Eq.~9 quantifies this effect and experiments in Tab.~1 of the main paper demonstrate the effectiveness of our solution. 

\begin{figure*}[t]
\centering
  \includegraphics[width=0.7\linewidth]{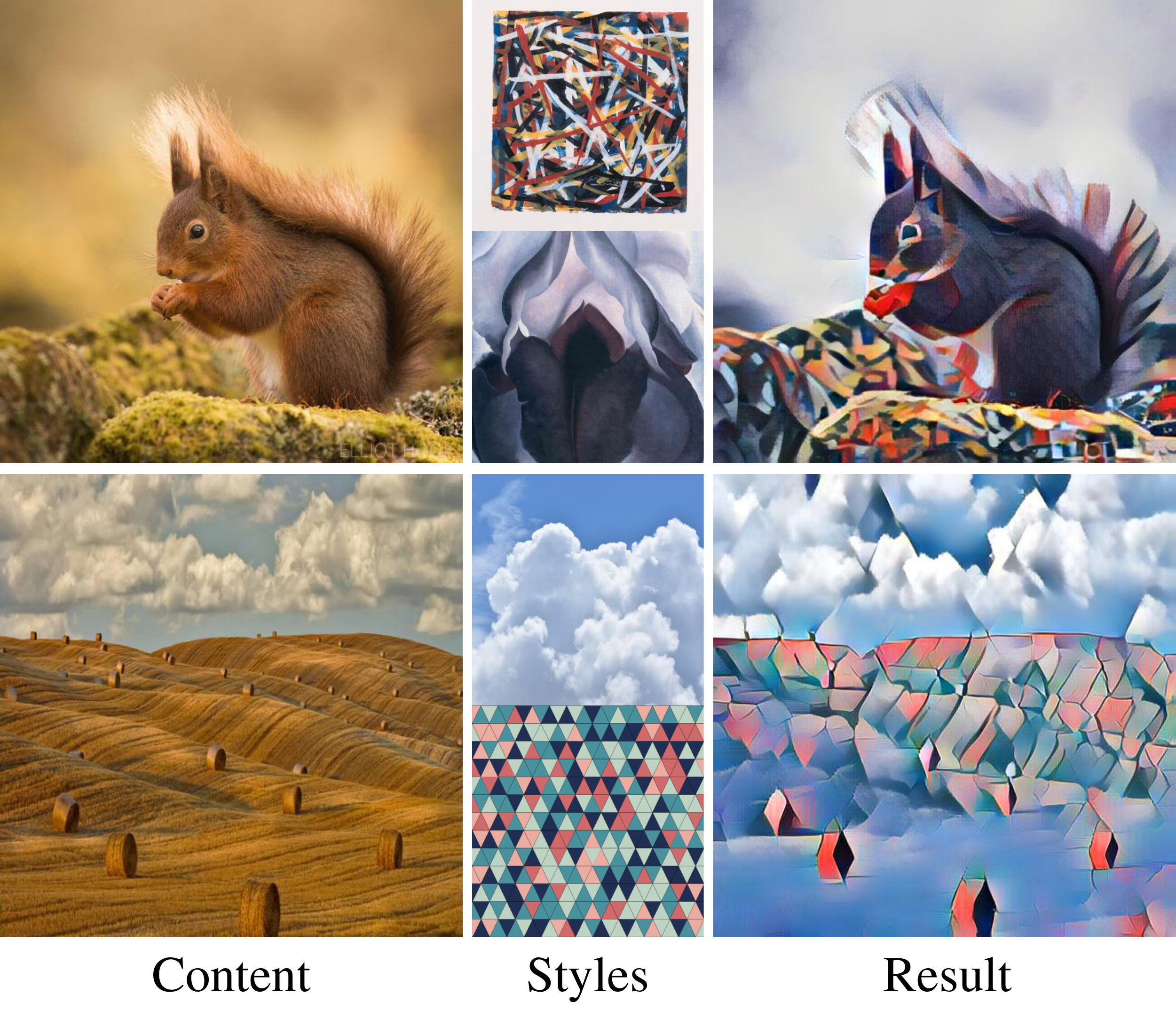}
  \caption{Results of multi-style transfer.}
  \label{fig:multistyle}
\end{figure*}

\begin{figure}[t]
\centering
  \includegraphics[width=\linewidth]{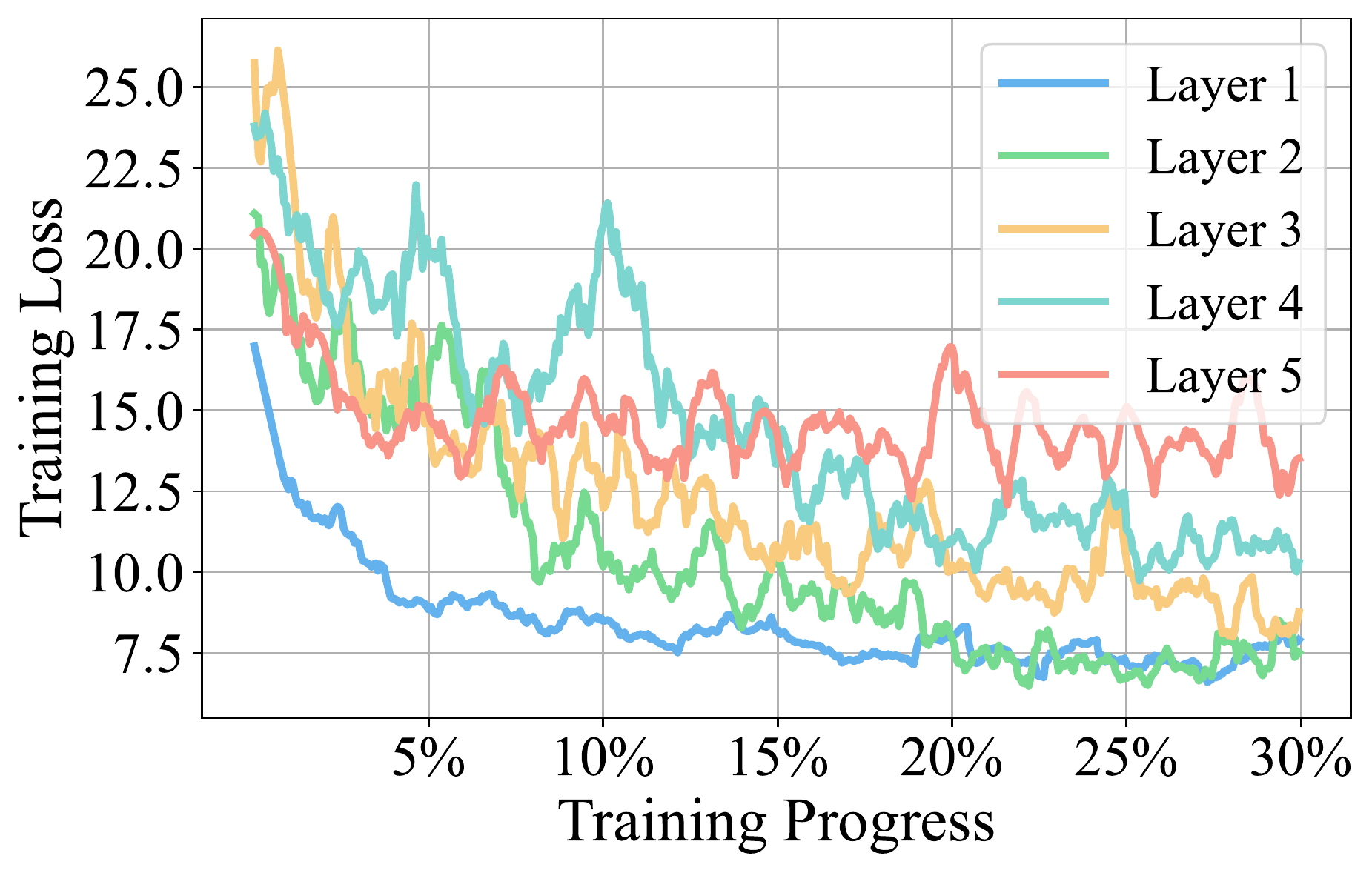}
  \caption{Fine-grained ablation studies on the number of layers used without parameter sharing to train a style transfer model.}
  \label{fig:converge}
\end{figure}

\textbf{Impact of Multiple Transformer Layers on Training Convergence:} 
One drawback of the vanilla Transformer model in style transfer is that the multi-layer structure can lead to difficult training convergence. 
As shown in Fig.~\ref{fig:converge}, with more layers adopted, the loss may converge more slowly, and it even fails in the 5-layer case. 
There seems to be a contradiction with the conclusion on the generative model focusing on StyleGAN~\cite{karras2020analyzing}: the model becomes more robust with more parameters. 
In fact, instead of generating new contents unconditionally in StyleGAN, style transfer aims to preserve contents and migrate style patterns at the same time. Stacking more layers in Transformer models may increase the complexity of the transfer function and tends to learn more abstract information. Thus, with more layers, it becomes harder to preserve original content structures during training.
Sharing parameters for different layers kills three birds with one stone: it makes a light-weight, easy-to-train, and easy-to-control model.

\subsection{More Extensions}
\noindent\textbf{Style Interpolation.} 
Our model also supports style interpolation by conducting linear interpolation to a couple of output features of our Style Transformer. 
Two examples are shown in Fig. \ref{fig:interpolate}.

\noindent\textbf{Multi-Style Transfer.} 
It is convenient for our method to achieve multi-style transfer by simply send features of multiple style images to the style encoder of our Master model.
Results are shown in Fig. \ref{fig:multistyle}.

\begin{figure*}[t]
\centering
  \includegraphics[height=\textheight]{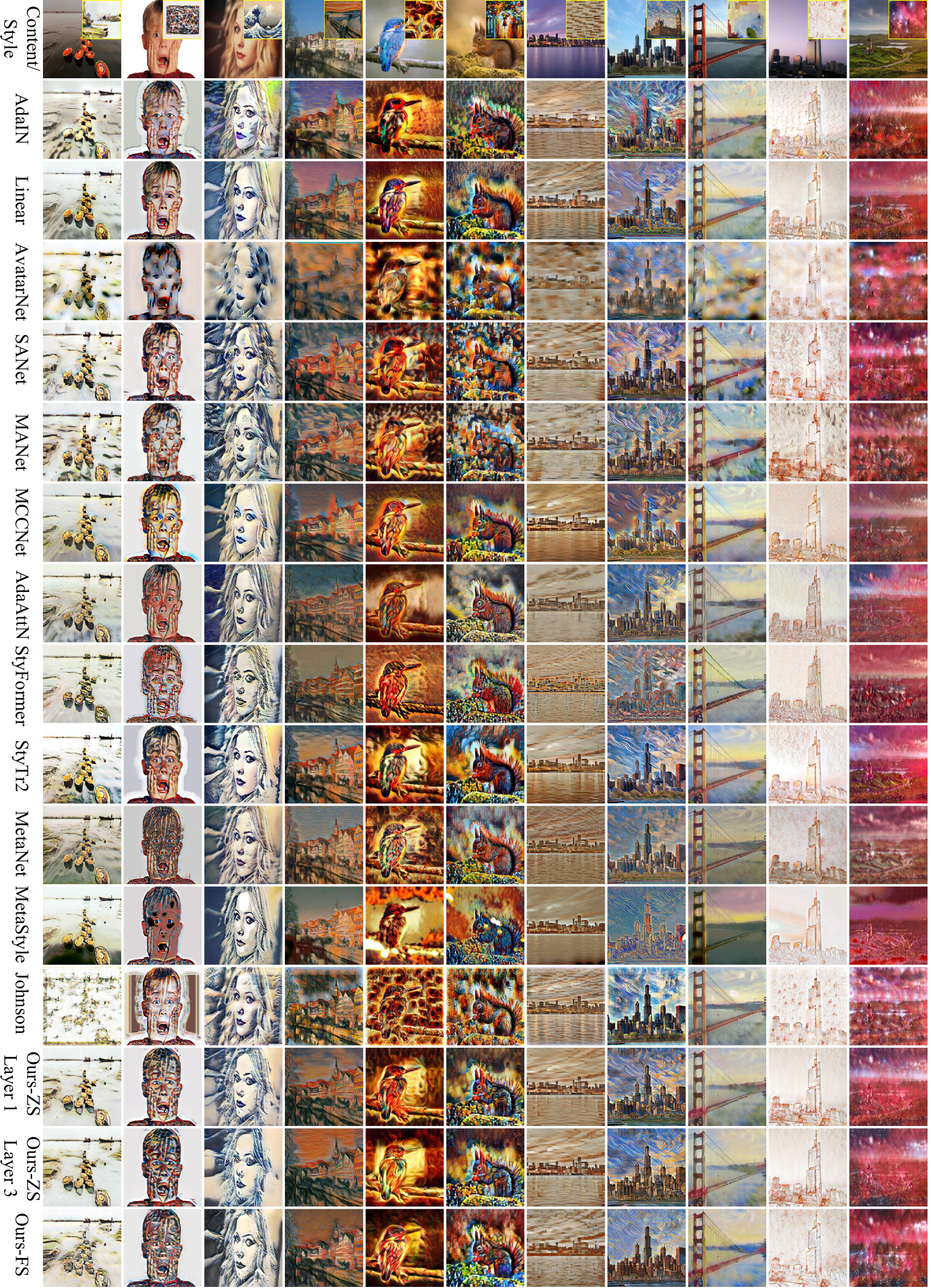}
  \caption{Full comparison results as a supplement to Fig. 4 in the main paper. Zoom in for better details.}
  \label{fig:compare_supp}
\end{figure*}

\begin{figure*}[t]
\centering
  \includegraphics[width=\textwidth]{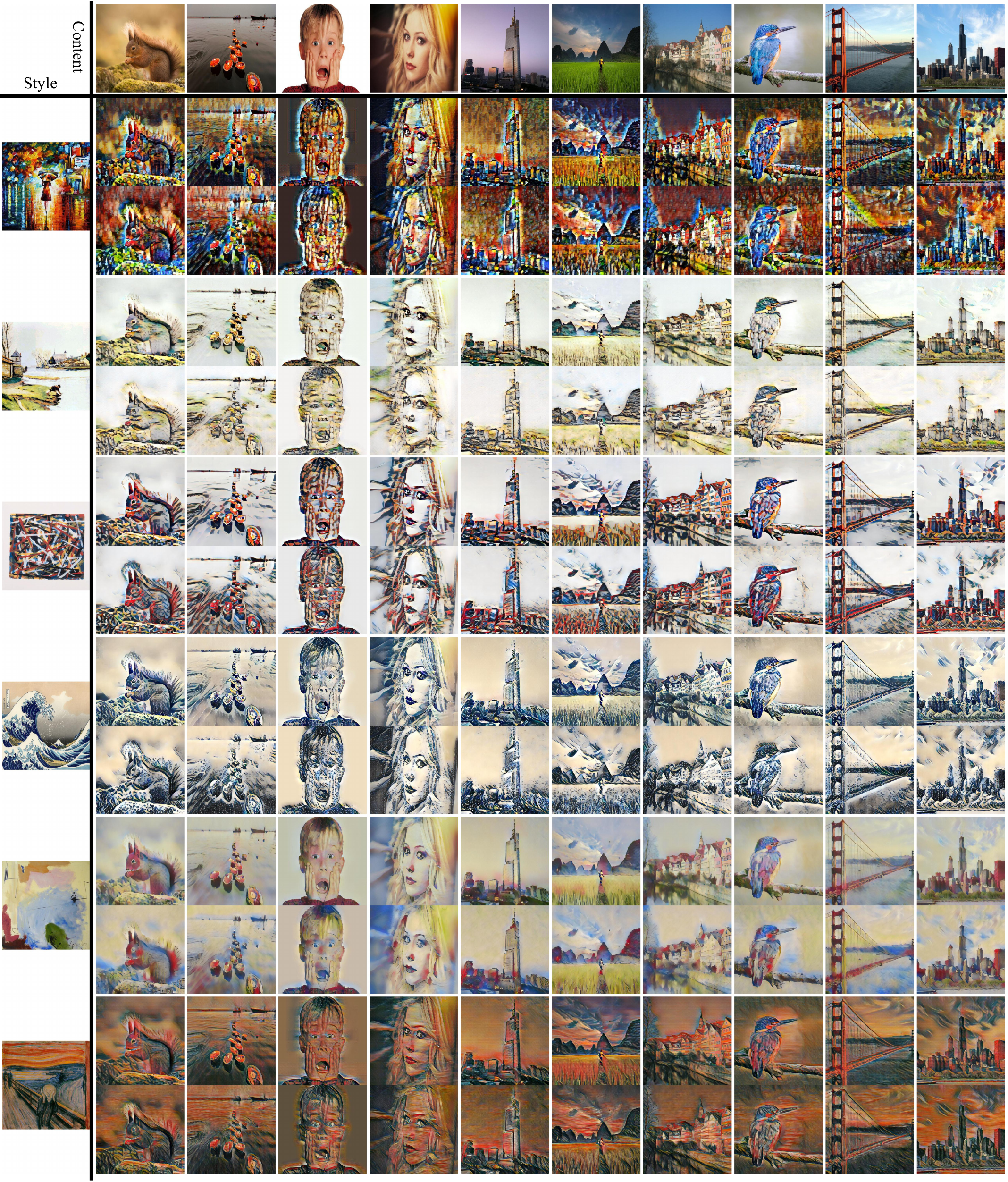}
  \caption{More content-style pairs. Upper and bottom images of each entry are results under zero-shot and few-shot settings respectively. Zoom in for better details.}
  \label{fig:pair}
\end{figure*}

\begin{figure*}[t]
\centering
  \includegraphics[width=\textwidth]{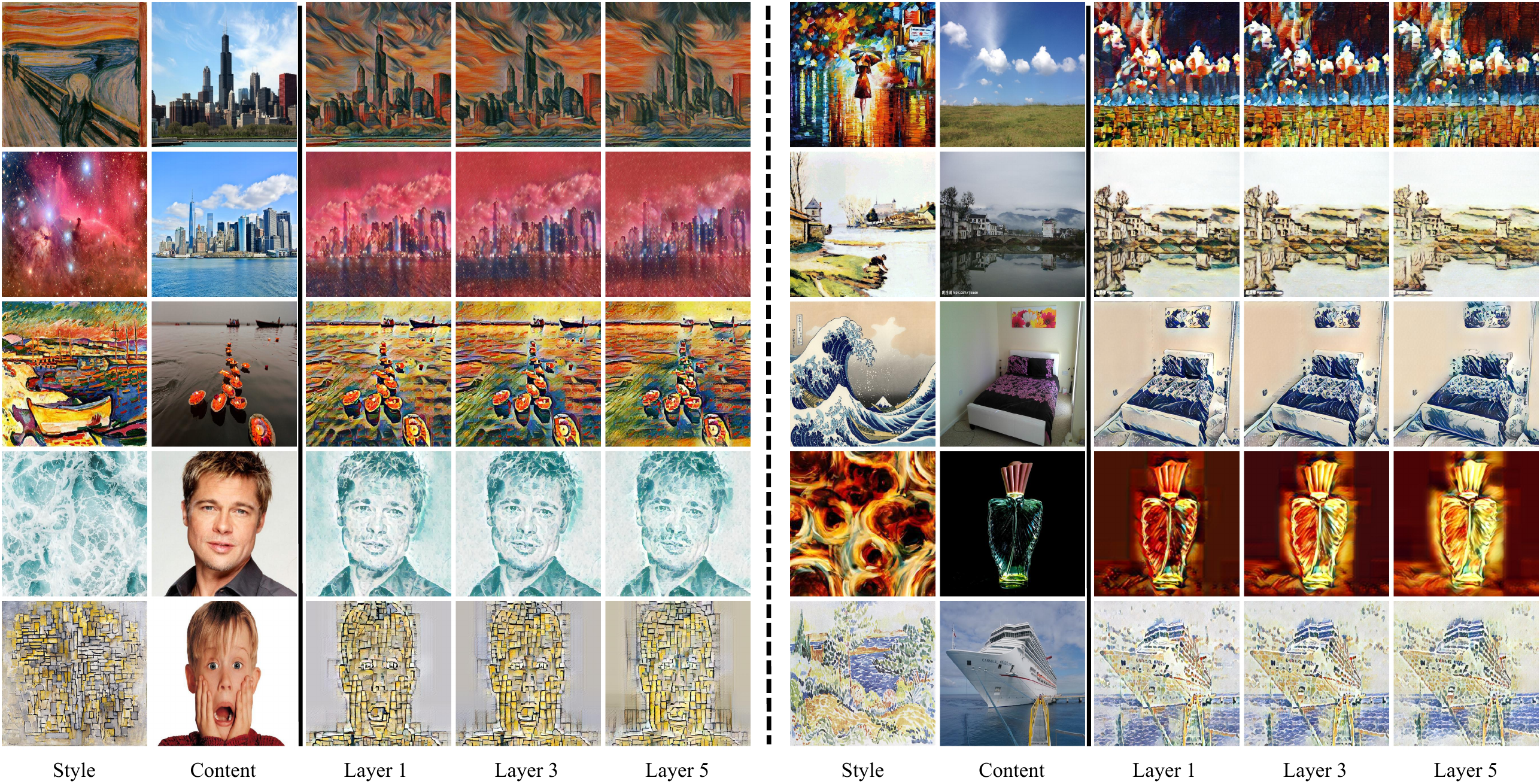}
  \caption{More controllable style transfer results by using different numbers of stacked Transformer layers in the test time.}
  \label{fig:control_supp}
\end{figure*}

\begin{figure*}[t]
\centering
  \includegraphics[width=\textwidth]{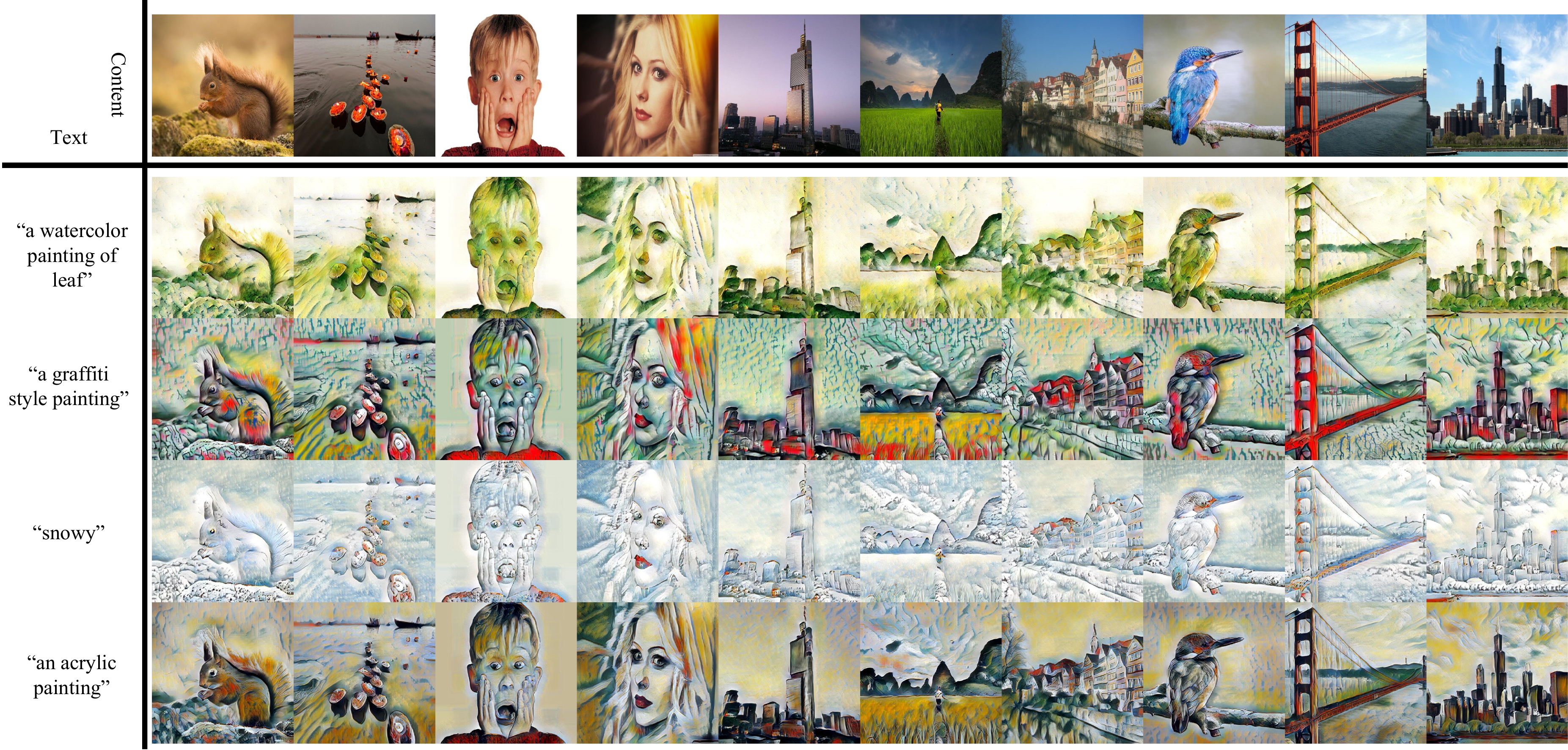}
  \caption{More content-text pairs for text-guided style transfer.}
  \label{fig:text_pair}
\end{figure*}

\begin{figure*}[t]
\centering
  \includegraphics[width=0.73\textwidth]{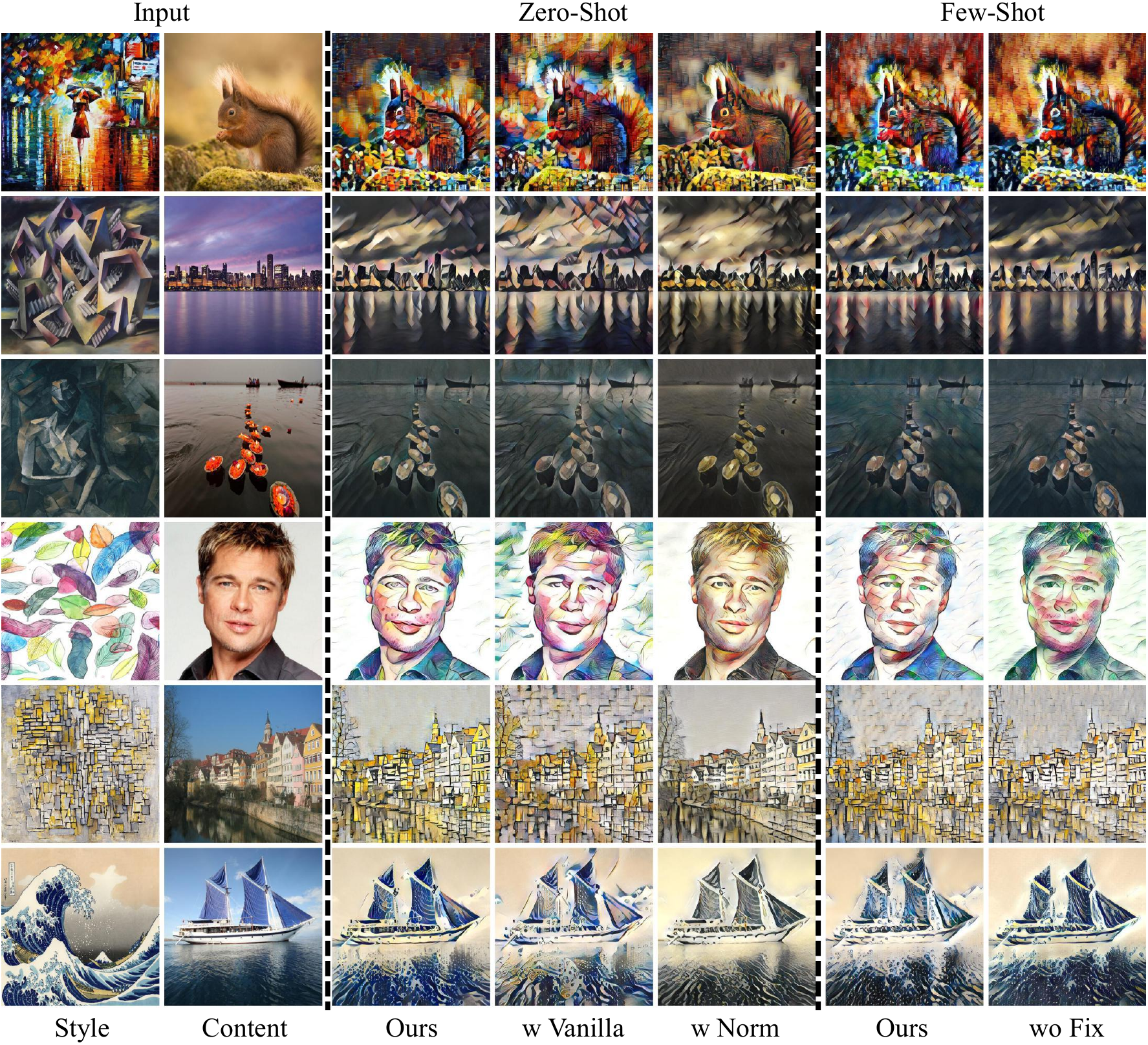}
  \caption{More ablation results as a supplement to Fig. 7 in the main paper. Zoom in for better details.}
  \label{fig:ablation_supp}
\end{figure*}

\begin{figure*}[t]
\centering
  \includegraphics[width=\textwidth]{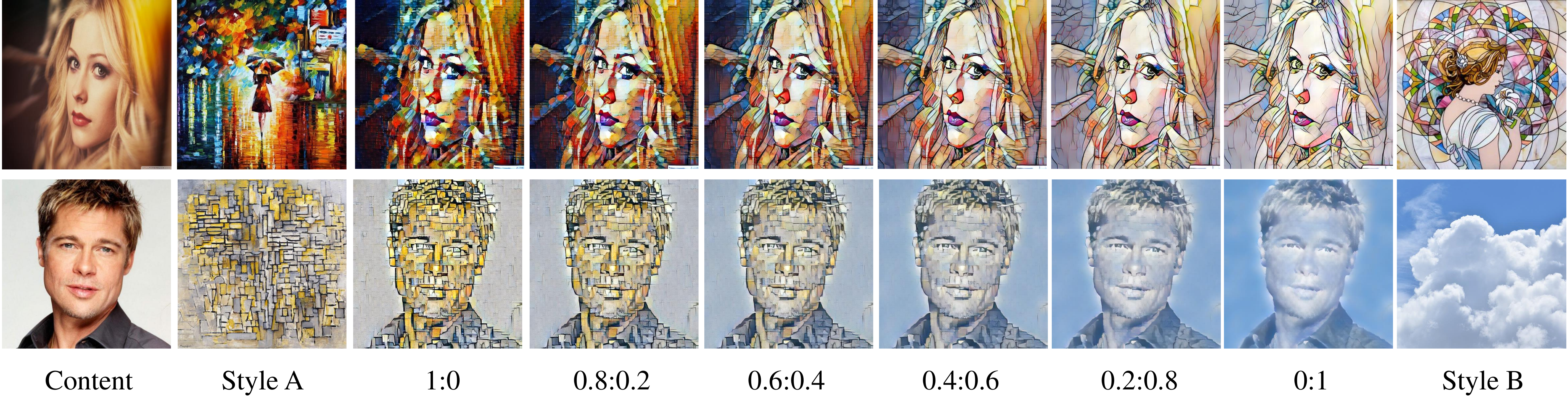}
  \caption{Two-style interpolation results. The content image and style images are shown on the two ends}
  \label{fig:interpolate}
\end{figure*}

\end{document}